\documentclass[sigconf]{acmart}
\setcopyright{acmcopyright}
\copyrightyear{2021}
\acmYear{2021}
\acmConference[WWW '21]{Proceedings of the Web Conference 2021}{April 19--23, 2021}{Ljubljana, Slovenia}
\acmBooktitle{Proceedings of the Web Conference 2021 (WWW '21), April 19--23, 2021, Ljubljana, Slovenia}
\acmPrice{}
\acmDOI{10.1145/3442381.3450112}
\acmISBN{978-1-4503-8312-7/21/04}

\usepackage[utf8]{inputenc} 
\usepackage[T1]{fontenc}    
\usepackage{hyperref}       
\usepackage{url}            
\usepackage{booktabs}       
\usepackage{amsfonts}       
\usepackage{nicefrac}       
\usepackage{microtype}      
\usepackage{graphicx}
\usepackage[ruled,vlined,linesnumbered]{algorithm2e}

\usepackage{pifont}

\usepackage{subfigure}

\usepackage{multirow}
\usepackage{amsmath}
\usepackage{url}
\usepackage{balance}
\usepackage{amsthm}
\usepackage{paralist}
 
\usepackage{fancyhdr,graphicx,amsmath,amssymb}
\usepackage[ruled,vlined]{algorithm2e}
\include{pythonlisting}
\usepackage{graphics}
\usepackage{hyperref}

\usepackage{xspace}

\newcommand{\cmarkb}{\ding{52}}

\usepackage{colortbl}
\usepackage{xcolor}
\usepackage[linesnumbered,ruled,vlined]{algorithm2e}

\SetCommentSty{mycommfont}
\SetKwInOut{KwInOut}{Require}                
\usepackage{enumitem}
\newtheorem{problem}{Problem}
\newcommand\blfootnote[1]{%
  \begingroup
  \renewcommand\thefootnote{}\footnote{#1}%
  \addtocounter{footnote}{-1}%
  \endgroup
}

\begin{document}

\title{Few-Shot Graph Learning for Molecular Property Prediction}

\author{Zhichun Guo$^{1}$, Chuxu Zhang$^{2*}$, Wenhao Yu$^{1}$\\
John Herr$^{1}$, Olaf Wiest$^{1}$, Meng Jiang$^{1}$, Nitesh V. Chawla$^{1*}$}
\affiliation{%
  \institution{{$^1$University of Notre Dame, IN, USA } \ \ {$^2$Brandeis University, MA, USA }\\
  }
}
\email{zguo5@nd.edu, chuxuzhang@brandeis.edu, {wyu1, jherr1, Olaf.G.Wiest.1, mjiang2, nchawla}@nd.edu}

\renewcommand{\shortauthors}{Zhichun Guo, Chuxu Zhang, Wenhao Yu, John Herr, Olaf Wiest, Meng Jiang, Nitesh V. Chawla}

\renewcommand{\shortauthors}{This work was accepted to The Web Conference (WWW) 2021.}

\begin{abstract}

The recent success of graph neural networks has significantly boosted molecular property prediction, advancing activities such as drug discovery. The existing deep neural network methods usually require large training dataset for each property, impairing their performance in cases (especially for new molecular properties) with a limited amount of experimental data, which are common in real situations. To this end, we propose Meta-MGNN, a novel model for few-shot molecular property prediction. Meta-MGNN applies molecular graph neural network to learn molecular representations and builds a meta-learning framework for model optimization. To exploit unlabeled molecular information and address task heterogeneity of different molecular properties, Meta-MGNN further incorporates molecular structures, attribute based self-supervised modules and self-attentive task weights into the former framework, strengthening the whole learning model. Extensive experiments on two public multi-property datasets demonstrate that Meta-MGNN outperforms a variety of state-of-the-art methods. 

\end{abstract}

\keywords{Molecular Property Prediction, Few-Shot Learning, Graph Learning}

\maketitle

{\fontsize{8pt}{8pt} \selectfont \textbf{ACM Reference Format:}\\
Zhichun Guo, Chuxu Zhang, Wenhao Yu, John Herr, Olaf Wiest, Meng Jiang, Nitesh V. Chawla. 2021. Few-Shot Graph Learning for Molecular Property Prediction. In \textit{Proceedings of The Web Conference 2021 (WWW '21), April 19-23, 2021,
Ljubljana, Slovenia.} ACM, New York, NY, USA, 9 pages. https://doi.org/10.1145/3442381.3450112}

\blfootnote{$*$ Corresponding authors}
\blfootnote{$\S$ Our code is avaiable at \url{https://github.com/zhichunguo/Meta-MGNN}}

\section{Introduction}

Drug discovery significantly benefits all human beings, especially for public health during this tough and special time caused by COVID-19~\cite{vamathevan2019applications}. Developing and discovering new drugs is a time, resource, and money consuming process. A key step is to test a large number of molecules for therapeutic activity through extensive biological studies~\cite{sliwoski2014computational}. Unfortunately, these discovered ones often fail to become the approved drug candidates for various reasons such as low activity or toxicity~\cite{waring2015analysis}. Researchers need to select a great number of similar molecules as potential candidates. To find the molecules which have the same efficacious property, these selected molecules need to be tested through a complex experimental process. After that, only a few or even no molecules will be remaining as possible drug candidates to be tested further for risk and pharmaceutical activity. Therefore, it is crucial to improve the effectiveness of filtering the most likely drug candidates before taking experiments via wet-lab experimentation, thus wasting less time and resources on molecules that are unlikely to proceed to the lead stage. This concept is generally described as "fail early-fail cheap".

Virtual screening is a widely used approach to screen out molecules likely to fail early, which avoids a large set of molecules to be investigated~\cite{riniker2013similarity, sliwoski2014computational}. Recent advances in deep learning have played an important role in virtual screening. These deep learning techniques have inspired novel approaches to a better understanding of molecules and their properties through molecular representation learning~\cite{hu2020strategies,xu2017seq2seq,gilmer2017neural,zhang2018seq3seq,zheng2019identifying}.
Deep neural networks learn more about specific molecular properties when they are fed with more instances during training. Thus, deep learning models require a large amount of training data to achieve desired capability and satisfactory performance~\cite{devlin2018bert}. However, it is common that there are only a few known molecules that share the same set of properties~\cite{altae2017low,waring2015analysis}. We analyzed the datasets in MoleculeNet~\cite{wu2018moleculenet}, a well-known benchmark for predicting molecular properties. We find that more than half of the properties only are shared by fewer than 100 molecules across several datasets. This is a case of the well-known problem of \emph{few-shot} available data, which seriously impairs the performances of current approaches. Therefore, it is essential to develop a deep neural model for predicting molecular properties effectively in few-shot scenarios. 

There are several challenges that need to be overcome to achieve this goal. Molecules can be considered as a heterogeneous structure where each atom connects to different neighboring atoms via different types of bonds. Previous work~\cite{wang2019smiles} represents molecules as \textit{SMILES} strings and leverages sequence models~~\cite{mikolov2013distributed,wang2019smiles} to learn molecular embedding. This approach is not able to capture information in each bond well~\cite{wu2020comprehensive}. This is because bonds in molecules not only represent connected relations between different atoms but also contain attributed information that characterizes the bond type such as single, double, or triple.
Thus, the first challenge is to design a deep neural network that can discover \emph{effective} molecular representations from \emph{few-shot} data. Because only a limited amount of labeled molecular property data are available, the second challenge is to exploit the useful \emph{unlabeled} information in molecule data and further develop an \emph{efficient} learning procedure to transfer the knowledge from other property prediction, so that the model can fast adapt to the novel (new) molecular properties with limited data. Moreover, different molecular properties could represent quite different molecular structures. Thus, their data should be treated differently in the knowledge transfer process. The third challenge is to distinguish the \emph{different} importance of molecular properties when performing the efficient learning procedure. 

To address the above challenges, we propose a novel model called Meta-MGNN for few-shot molecular property prediction. First, we leverage graph neural network with the pre-training process to fuse heterogeneous molecular graph information as molecular embedding. Then, we develop a meta-learning framework to transfer knowledge from different property prediction tasks and obtain a well-initialized model which could be fast adapted to a new molecular property with limited data. In order to exploit and capture unlabeled information in molecule data, we design a self-supervised module which consists of a bond reconstruction loss and an atom type prediction loss, accompanied by the main property prediction loss. Moreover, considering different property prediction tasks contribute differently to the few-shot learner, we further introduce a self-attentive task weight to measure their importance. Both self-supervised module and self-attentive task weight are incorporated into the meta-learning procedure for strengthening the model. 


\noindent\textbf{Contributions.} To summarize, the main contributions of this work are as follows:
\begin{itemize}[leftmargin=*]
\item We formulate the molecular property prediction as a few-shot learning problem, which exploits the rich information in various properties to address the lack of laboratory data problem for each individual property. 
\item To deal with the few-shot challenge, we propose a novel model called Meta-MGNN by exploring graph neural network, self-supervised learning, and task weight aware meta-learning. 
\item We conduct extensive experiments on two public datasets and the evaluation results demonstrate the superior performance of Meta-MGNN over state-of-the-art methods. The effectiveness of each model component is also verified.  
\end{itemize}


\begin{figure*}[t]
    \centering
    {\includegraphics[width=1.0\textwidth]{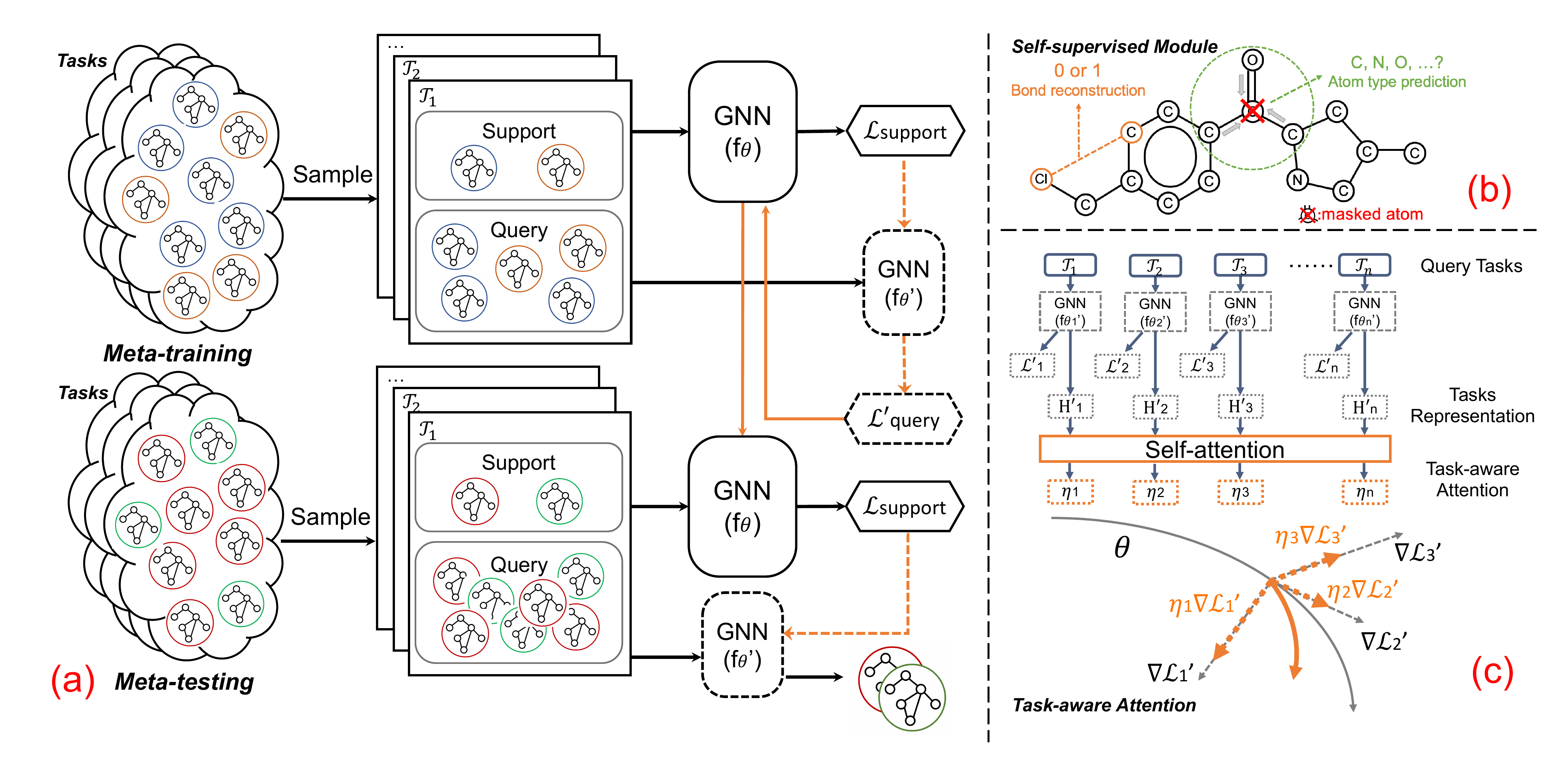}}
    \vspace{-0.1in}
    \caption{(a) The overall framework of Meta-MGNN: It first samples a batch of training tasks. For each task, there are a few data examples in the support set. These examples are fed into a GNN parameterized by $\theta$. Then the support loss $\mathcal{L}_{support}$ is calculated and utilized to update the GNN parameters to $\theta'$. Next, the examples in the corresponding query set are fed into the GNN parameterized by $\theta'$ and calculate the loss $\mathcal{L}_{query}'$ for this task. The same process repeats for other training tasks. Later, we compute the summation of  $\mathcal{L}_{query}'$ over all sampled tasks and use it to further update the GNN parameters for testing. (b) Self-supervised module: It includes bond reconstruction and atom type prediction. The orange part shows that we sample two atoms and use GNN to predict if there is a bond between them. The green part shows that we mask several atoms randomly and use GNN to predict their types. (c) Task-aware attention: It calculates the average of all the molecular embedding from the query set of the same task to represent this task. With the embedding of each task, we design a self-attentive layer to compute the weight of each task, then incorporate it into a meta-training process for updating model parameters $\theta$.}
\label{fig:mole_repres}
\end{figure*}

\section{Related Work}
In this section, we review existing work including graph neural network, few-shot learning, and molecular property prediction.

\vspace{0.05in}
\noindent\textbf{Graph Neural Network (GNN). } GNNs have gained increasing popularity due to its
capability of modeling graph-structured data~\cite{hamilton2017inductive,velivckovic2018graph,zhang2019heterogeneous}. 
Typically, a GNN model uses a neighborhood aggregation function to iteratively update the representation of a node by aggregating representations of its neighboring nodes and edges.
GNNs have showed attractive performance 
in various applications, such as recommendation systems~\cite{fan2019graph,song2019session}, behavior modeling~\cite{yu2020identifying}, and anomaly detection~\cite{zhao2020early}.
Molecular property prediction is also a popular  application of GNNs since a molecule could
be represented as a topological graph by treating atoms as nodes, and bonds as edges~\cite{gilmer2017neural,mansimov2019molecular,lu2019molecular,guo2020graseq}. We will elaborate them in the next paragraph.

\vspace{0.05in}
\noindent\textbf{Molecular Property Prediction. } 
Methods can be categorized into two main groups based on the input molecular type: (1) molecular graph, and (2) simplified molecular-input line-entry (\textit{SMILES})~\cite{weininger1989smiles}. 
For the first group of methods, each molecule is represented as a graph associated with different atom nodes interconnected by bond edges. One typical way is to employ graph neural networks to learn molecular representations~\cite{gilmer2017neural,hu2020strategies,mansimov2019molecular,lu2019molecular,guo2020graseq}. For example, Lu et al.~\cite{lu2019molecular} proposed a novel hierarchical GNN. It includes an embedding layer, a Radial Basis Function layer, and an interaction layer to learn molecular representations from different levels. Hu et al.~\cite{hu2020strategies} proposed several novel pre-training strategies to pre-train GNNs at the level of individual nodes and the entire graph to lean local and global molecular representations simultaneously.
For the second type of representation, \textit{SMILES} is a sequence notation for describing the structure of molecules. Researchers take molecules as sequences and adopt language models to learn their representations~\cite{zhang2018seq3seq,wang2019smiles,zheng2019identifying}. For example, Zhang et al.~\cite{zhang2018seq3seq} proposed a semi-supervised Seq2Seq fingerprint model which contains three ends of one input, one supervised output, and one unsupervised output. Zheng et al.~\cite{zheng2019identifying} presented a new model to study structure-property relationships through a self-attentive linear notation syntax analysis. 
Guo et al.~\cite{guo2020graseq} proposed a novel graph and sequence fusion learning model to capture information both from the molecular graph structure and \textit{SMILES}. Here, we take each molecule as a graph as it preserves the molecular inner structure better, and employ graph neural networks to learn their representations.

\vspace{0.05in}
\noindent\textbf{Few-shot Learning. } 
Successes of few-shot learning have been accomplished in various application domains such as computer vision~\cite{finn2017model,kim2019edge} and graph learning~\cite{yao2019learning,yao2020graph,garcia2018few,zhang2020few,zhang2020few2}
. There are two notable types of few-shot learning approaches: (1) \textit{metric-based learning} and (2) \textit{gradient-based learning}. The former learns a generative metric to compare and match few-examples ~\cite{vinyals2016matching,sung2018learning,bertinetto2018meta}. Vinyals et al.~\cite{vinyals2016matching} proposed a novel matching metric, named Matching Nets, to match unlabeled examples to the class of few-shot labeled examples. Sung et al.~\cite{sung2018learning} proposed relation network which learns a deep distance metric to compute relation scores of different images and further classify images. The latter aims to employ a specific meta-learner to learn well-initialized parameters of the base model for different tasks~\cite{finn2017model, lee2018gradient, zhang2018metagan}. For instance, Finn et al.~\cite{finn2017model} proposed MAML which designs this kind of meta-learner to effectively initialize a base-learner that could be fast adapted to new tasks. 
In this work, our few-shot learning strategy is gradient-based learning.

\section{Preliminary}
In this section, we first define the few-shot molecular property prediction problem, then present the details of using graph neural network (GNN) for learning molecular representations. 
\subsection{Problem Definition}
Let $G = (\mathcal{V}, \mathcal{E})$ denote a molecular graph where $\mathcal{V}$ is the set of nodes and $\mathcal{E} \subseteq \mathcal{V} \times \mathcal{V}$ is the set of edges. 
Particularly, a node in a molecular graph represents a chemical atom and an edge represents a chemical bond between two atoms. Given a set of molecular graphs $\mathcal{G} = \{G_1, \cdots, G_N\}$ and their labels $\mathcal{Y} = \{y_1, \cdots, y_N\}$, the goal of molecular property prediction is to learn a molecular representation vector for predicting its label (i.e., molecular property) of each $G_i \in \mathcal{G}$, i.e., to learn a mapping function $f_\theta: \mathcal{G} \rightarrow \mathcal{Y}$.

Unlike previous studies where there are enough examples for each new property prediction task, this work considers a more practical scenario that only few-shot samples are given. Specifically, we aim to develop a classifier which can be fast adapted to predict new molecular properties that are \textit{unseen} during the training process, given only a few samples of these new properties. Formally, the problem is defined as follows. 
\begin{problem}
{\bf Few-Shot Molecular Property Prediction}
Given molecular properties $\mathcal{Y} = \{y_1, \cdots, y_N\}$ and their corresponding few-shot molecular graph sets $\{{\mathcal{G}_1} \in y_1, \cdots, {\mathcal{G}_N} \in y_N\}$ (training data), the task is to design a machine learning model to predict molecular graphs of new properties that only have few-shot examples (test data). 
\end{problem}

\subsection{Molecular Graph Neural Network}
By viewing molecular structure as graph data (i.e., molecular graph), recent deep learning methods for graphs, such as graph neural networks (GNNs)~\cite{wu2020comprehensive,zhang2019heterogeneous}, can be utilized to learn molecular representations which are fed to downstream machine learning models for molecular property prediction~\cite{liu2019chemi}. In this section, we will present the details of employing GNNs to obtain molecular representations.

A GNN model is able to utilize both graph structure and node/edge features information to learn a representation vector $\textbf{h}_v$ for each node $v \in \mathcal{V}$. Specifically, a GNN model uses a neighborhood aggregation function to iteratively update the representation of a node by aggregating representations of its neighboring nodes and edges. After $l$ iterations, a node representation $\textbf{h}_v^{(l)}$ is able to capture the information within its $l$-hop neighborhoods. 
In a molecular graph, each node represents an atom and each edge represents a chemical bond between two atoms. As the input layer of GNN, we first initialize representations of both nodes and edges using their attributes in molecular graph. The node attributes include atom number (AN) and chirality tag (CT), and edge attributes include bond type (BT) and bond direction (BD). Formally, we initialize node representation as $\textbf{h}_v^{(0)} = \textbf{v}_{AN} \oplus \textbf{v}_{CT}$ and edge representation as  $\textbf{h}_e^{(0)} = \textbf{e}_{BT} \oplus \textbf{e}_{BD}$, where $\textbf{v}$ and $\textbf{e}$ denote node/edge attributes and $\oplus$ is concatenation operator. Then, the node representation $\textbf{h}_v^{(l)}$ at the $l$-th layer of GNN is formulated as:
\begin{align}
    \textbf{h}^{(l)}_{\mathcal{N}(v)} = \textsc{Agg}_l( \{\textbf{h}^{(l-1)}_u: \forall u \in \mathcal{N}(v) \}, \{ \textbf{h}^{(l-1)}_e: e=(v, u)\}),
\end{align}
\begin{align}
    \textbf{h}^{(l)}_v = \sigma (\textbf{W}^{(l)} \cdot \textsc{Concat}(\textbf{h}^{(l-1)}_v, \textbf{h}^{(l)}_{\mathcal{N}(v)})),    
\end{align}
where $\mathcal{N}(v)$ is the neighbor set of $v$, $\sigma(\cdot)$ is a non-linear activation function (e.g., LeakyReLU). $\textsc{Agg}(\cdot)$ is an aggregating function. A number of architectures for $\textsc{Agg}(\cdot)$ have been proposed in recent years such as graph convolutional neural network (GCN)~\cite{kipf2017semi} and graph attention network (GAT)~\cite{velivckovic2018graph}. Here, we use graph isomorphism network (GIN)~\cite{xu2019powerful}, which has demonstrated state-of-the-art performance on a variety of benchmark tasks. After that, we can learn the representation of each node in molecular graph: $\textbf{h}_v = \textbf{h}^{(l)}_v/ || \textbf{h}^{(l)}_v||_2$. To obtain the graph-level representation $\textbf{h}_G$ for a molecular graph, we calculate the average node embeddings at the final layer: 
\begin{align}
    \textbf{h}_G = \textsc{Mean}(\{\textbf{h}_{v}^{(l)}: v \in \mathcal{V} \}),
    \label{eq:graph-level}
\end{align}
The graph-level molecular representation $\textbf{h}_G$ can be further fed into a classifier (e.g., a multi-layer perception) for  molecular property prediction, as we will present in the next section.

\vspace{0.05in}
\noindent{\textbf{Pre-trained Molecular Graph Neural Network.}} Pre-trained models have been widely used in natural language processing, computer vision, and graph analysis in recent years~\cite{devlin2018bert,hu2020strategies}. In general, pre-training allows a model to learn universal representations, provides a better parameter initialization, and avoids overfitting on downstream tasks with small training data. Models with pre-training have been demonstrated to obtain superior performance than models without that. Therefore, we are motivated to leverage the recent pre-trained graph neural network technique (PreGNN)~\cite{hu2020strategies} to obtain parameter initialization of molecular graph neural network. 

\section{Meta-MGNN}
In this section, we present the details of proposed the Meta-MGNN for few-shot molecular property prediction. Meta-MGNN is built on MGNN and employs a meta-learning framework for model initialization and adaption. Molecular structure and feature based self-supervised module and self-attentive task weight are further incorporated into the former framework for model enhancement. 
\subsection{Meta-learning Setup}
We build the meta-learning framework based on MAML~\cite{finn2017model}.  Given the model $f_{\theta}$ with learnable parameters $\theta$ that maps molecular graph to specific properties such as toxicity, i.e., $f_{\theta}: \mathcal{G} \rightarrow \mathcal{Y}$. In meta-learning, the model is expected to adapt to a number of different tasks, i.e., predicting different kinds of molecular properties. Particular, in the $k$-shot meta-learning, for each task $\mathcal{T}_\tau$ sampled from distribution $p(\mathcal{T})$, the model is trained using only $k$ data samples and further tested on remaining data samples of $\mathcal{T}_\tau$. 
In this setting, we refer to the corresponding training and test sets of each task as \textit{support} set and \textit{query} set, denoted as  $\mathcal{T}_\tau = \{ \mathcal{G}_\tau, \mathcal{Y}_\tau, {\mathcal{G}}_\tau^{\prime}, {\mathcal{Y}}_\tau^{\prime}\}$, where $\mathcal{G}_\tau, \mathcal{Y}_\tau$ are \textit{support} sets of input molecular graphs and property labels, and $\mathcal{G}_\tau^\prime, \mathcal{Y}_\tau^\prime$ are \textit{query} sets of input molecular graphs and property labels. During meta-training, the model $f_{\theta}$ is first updated to  task-specific model using \textit{support} set of each task, then further optimized to task-agnostic model using prediction loss over the \textit{query} set of all tasks in training data. After sufficient training, the learned model can be further utilized to predict new tasks (new molecular properties) with only $k$ data samples as support set, which is called meta-testing. To avoid data overlapping, data of tasks used for meta-testing are held out during meta-training. The whole framework is illustrated in Figure 1(a). 



\subsection{Meta-training}
In meta-training, the goal
is to obtain well initialized model parameters $\theta$ that can be generally applicable to different tasks, and explicitly encourage the initialized parameters to perform well after a small number of gradient descent updates on a new task with few-shot data. 
When adapting to a task $\mathcal{T}_\tau$, we begin with feeding the \textit{support} set to the model and calculate the loss $\mathcal{L}_{\mathcal{T}_\tau}$ to update parameters $\theta$ to $\theta^{\prime}_\tau$ through gradient descent:
\begin{equation}
    \theta^{\prime}_\tau = \theta - \alpha \nabla_{\theta}\mathcal{L}_{\mathcal{T}_\tau} (\theta),
    \label{eq:update_task}
\end{equation}
where $\alpha$ is the step size. It should be noted that Eq.(\ref{eq:update_task}) only shows one-step
gradient update while we can take multiple-steps gradient update in practice.

\subsubsection{\textbf{Loss Function.}}
Typically, the above loss $\mathcal{L}_{\mathcal{T}_\tau}$ is calculated by the supervised signals from downstream tasks~\cite{finn2017model,zhou2019meta}, i.e., molecular property labels in this study. 
However, simply using supervised signals maybe not effective since only a few samples are given for each task. In addition, the complexity of molecules inherently bring useful unlabeled information in both structure and attribute. 
Therefore, to enhance the above meta-training process, we propose to exploit and leverage unlabeled information in molecular graphs. In particular, we design a self-supervised module which consists of a \textit{bond reconstruction loss} and an \textit{atom type prediction loss}, accompanying with the property prediction loss. 

\vspace{0.05in}
\noindent{\textbf{Molecular Property Prediction Loss.}} To predict molecular property, we introduce a multi-layer perception (MLP) on top of the graph-level molecular representation $\textbf{h}$ ( Eq.(\ref{eq:graph-level})), i.e., $\hat{y} = \textsc{MLP} (\textbf{h})$. The loss of prediction is defined as the \textit{cross entropy} loss between the predicted labels and ground-truth labels: 
\begin{equation}
    \mathcal{L}_{label}(\theta) = - \frac{1}{k} \sum_{i=1}^{k} \textsc{CrossEntropy}(y_i, \hat{y}_i)
\end{equation}
where $k$ is the number of data samples. 
\begin{algorithm}[t]
\DontPrintSemicolon
\SetAlgoLined
\KwInOut{\{$\mathcal{G}_\tau, \mathcal{Y}_\tau$\}: support\ data\ ; $\{{\mathcal{G}}_\tau^{\prime}, {\mathcal{Y}}_\tau^{\prime}$\}: query\ data; \
$\alpha$, $\beta$: step sizes (i.e., learning\ rates)}
$\theta$ $\leftarrow$ Pre-trained\ by\ PreGNN~\cite{hu2020strategies}\ \\
\While{\textbf{not done}}{
    Sample batch of tasks $\mathcal{T}_{\tau} \sim p(\mathcal{T})$ \\
    \For{\textbf{all} $\mathcal{T}_{\tau}$}{
        Sample $k$ examples $\{G_{\tau1}, G_{\tau2}, \cdots, G_{\tau k}\} \in \mathcal{G}_\tau$ \\
        \For{i=1 to k}{
            $y_{\tau i}, \textbf{h}_{\tau i} = \mathrm{GNN}(G_{\tau i},\ \theta)$ \\
        } 
        $\textbf{H}_\tau = $ \textsc{Mean}\ ($\textbf{h}_{\tau1}, \textbf{h}_{\tau2}, \cdots, \textbf{h}_{\tau k}$)\\
        $\mathcal{L}_\tau \leftarrow$ Eq. (\ref{eq:loss-tau})\ with\ $\{y_{\tau1}, y_{\tau2}, \cdots, y_{\tau k}\}$ \\
        $\theta_\tau' = \theta\ -\ \alpha \nabla \mathcal{L}_\tau$ \\
        Sample n examples $\{G_{\tau1}',G_{\tau2}',...G_{\tau n}'\} \in \mathcal{G}_\tau^{\prime}$\\
        \For{j = 1 to n}{
            $y_{\tau j}', \textbf{h}_{\tau j}' = \mathrm{GNN}(G_{\tau j}',\ \theta_\tau')$ \\
        } 
        $\mathcal{L}_\tau' \leftarrow$ Eq. (\ref{eq:loss-tau})\ with\ $\{y_{\tau1}', y_{\tau2}', \cdots, y_{\tau n}'\}$\\
    }
    $\{\eta(\mathcal{T}_{1}), \cdots, \eta(\mathcal{T}_{t})\} \leftarrow$ Eq. (\ref{eq:attn})\ with\ $\{\textbf{H}_1, \cdots, \textbf{H}_t\}$ \\
    $\theta \leftarrow \theta - \beta \nabla_\theta \sum_{\mathcal{T}_\tau \sim p(\mathcal{T})} \eta(\mathcal{T}_{i}) \cdot \mathcal{L}_i'$ \\
}
\caption{Meta-MGNN}
\end{algorithm}

\vspace{0.05in}
\noindent{\textbf{Bond Reconstruction Loss.}} To perform bond reconstruction in molecular graphs, 
we first sample a set of positive edges (existing bonds) in the molecular graph, then sample a set of negative edges (non-existing bonds) by choosing node pairs that do not have an edge in the original molecular graph. We denote $\mathcal{E}_{s}$ as the union set of sampled positive edges and negative edges. In practice, we set $|\mathcal{E}_{s}|$ = 10 including 5 positive samples and 5 negative samples. The bond reconstruction score is computed by the inner product of embeddings between the sampled pair of nodes, i.e., $\hat{e}_{uv} = \textbf{h}_v^{\top} \cdot \textbf{h}_u $. 
The bond reconstruction loss is defined as the binary cross entropy loss between the predicted bonds and ground-truth bonds: 
\begin{equation}
    \mathcal{L}_{edge}(\theta) = - \frac{1}{|\mathcal{E}_{s}|} \sum_{e_{uv} \in \mathcal{E}_s} \textsc{BinaryCrossEntropy}(e_{uv}, \hat{e}_{uv})
\end{equation}

\vspace{0.02in}
\noindent{\textbf{Atom Type Prediction Loss.}} In a molecule, different atoms are connected in a certain way (e.g., carbon–carbon bond, carbon–oxygen bond), leading to different molecular structure. The atom type determines how a node in the molecular graph connects with neighboring nodes. Thus, we utilize the contextual sub-graph of a node (atom) to predict its type. Specifically, we first sample a set of nodes in a molecular graph, denoted as $\mathcal{V}_{ct} \subseteq \mathcal{V}$. For each node $v$ in $\mathcal{V}_{ct}$, the contextual sub-graph is defined as its neighbors within $l$-hops, i.e., $G_{sub} = (\mathcal{U}_{sub}, \mathcal{E}_{sub})$ where $\mathcal{U}_{sub} = \{v\} \cup \mathcal{N}_l(v) $, $\mathcal{E}_{sub} \subseteq \mathcal{U}_{sub} \times \mathcal{U}_{sub}$, and $\mathcal{N}_l(v)$ represents the the set of neighboring nodes of node $v$. In practice, we choose $\mathcal{V}_{ct} = 15\%\ $nodes in the graph and $l=1$. Later, we use a a multi-layer perception (MLP) on the top of mean pooling of all nodes in the contextual sub-graph excluding the central node and the atom type prediction loss is formulated as the cross entropy loss between predicted node type and ground-truth node type: 
\begin{equation}
    \hat{v}_i = \textsc{Mlp}~(\textsc{Mean}(\{\textbf{h}_{u}: u \in \mathcal{N}_l(v)\})),
\end{equation}
\begin{equation}
    \mathcal{L}_{node}(\theta) = - \frac{1}{|\mathcal{V}_{c}|} \sum_{i=1}^{|\mathcal{V}_{c}|} \textsc{CrossEntropy}(v_i, \hat{v}_i),
\end{equation}
The self-supervised module (of both bond reconstruction and atom type prediction) is illustrated in Figure 1(b). 

\vspace{0.05in}
\noindent{\textbf{Joint Loss.}} The loss for task $\mathcal{T}_\tau$ in the meta-training process is formulated as the summation over the above three losses,
\begin{equation}
    \mathcal{L}_{\mathcal{T}_\tau}(\theta) = \mathcal{L}_{node}(\theta) + \lambda_1 \mathcal{L}_{edge}(\theta) + \lambda_2 \mathcal{L}_{label}(\theta)
    \label{eq:loss-tau}
\end{equation}
where $\lambda_1$ and $\lambda_2$ are trade-off parameters that control the importance of different losses. In practice, we set $\lambda_1 = \lambda_2 = 0.1$.

\subsubsection{\textbf{Task-aware Attention.}}
With the new model parameter $\theta'_{\tau}$ obtained from the support set data of task $\mathcal{T}_{\tau}$, i.e., $\theta^{\prime}_\tau = \theta - \alpha \nabla_{\theta}\mathcal{L}_{\mathcal{T}_\tau} (\theta)$, the model is further updated as follows:
\begin{equation}
    \theta \leftarrow \theta - \beta \nabla_\theta \sum_{{\mathcal{T}_\tau} \sim p(\mathcal{T})} \eta(\mathcal{T}_{\tau}) \cdot \mathcal{L}_{\mathcal{T}_\tau}^{\prime}(\theta'_{\tau}),
\end{equation}
where $\beta$ is meta-learning rate, $\mathcal{L}_{\mathcal{T}_\tau}^{\prime}$ is the joint loss over \textit{query} set of $\mathcal{T}_{\tau}$. 
In other words, the model parameters $\theta$ are further updated over losses of all sampled tasks through gradient descent.
The traditional meta-learning methods (e.g., MAML~\cite{finn2017model}) treat each task with the same weight when optimizing the meta-leaner (i.e., $\eta(\mathcal{T}_{\tau})$ are same for all tasks), which cannot reflect how important of different property prediction tasks are. Therefore, considering different property prediction tasks contribute differently to the meta-learner optimization, we further introduce a self-attentive weight to measure task importance. Particular, we use self-attentive mechanism~\cite{lin2017structured} to calculate importance of each task:
\begin{equation}
    \eta({\mathcal{T}_\tau}) = \frac{\exp(\textsc{Mlp}(\textbf{H}_{\mathcal{T}_{\tau}}))}{\sum_{\mathcal{T}_{\tau'} \in \mathcal{T}} \exp(\textsc{Mlp}(\textbf{H}_{\mathcal{T}_{\tau'}}))},~ \textbf{H}_{\mathcal{T}_{\tau}} = \textsc{Mean}(\{\textbf{h}_{\mathcal{T}_{\tau}, i}\}_{i=1}^{k}).
    \label{eq:attn}
\end{equation}
where $\mathcal{T}$ is the set of all tasks, $\textbf{H}_{\mathcal{T}_{\tau}}$ denotes the task embedding which is computed by averaging all molecular embeddings of $\mathcal{T}_{\tau}$. Figure 1(c) illustrates the details of self-attentive task weight computation. The meta-training process is described in Algorithm 1. 

\subsubsection{\textbf{Meta-testing.}}
During meta-testing, we first utilize the few-shot \textit{support} set of new tasks to update parameters $\theta$ of Meta-MGNN via one or a small number of gradient descent steps using Eq. (\ref{eq:update_task}), then evaluate performance in  \textit{query} set.

\section{Experiments}
\begin{table}[t]
\centering
\caption{Details of atom and bond features.}
\vspace{-0.1in}
\setlength{\tabcolsep}{3mm}{
{\scalebox{0.90}{%
\begin{tabular}{c|c}
\toprule
\textbf{\# Atom Type} & 118\\
\midrule
\multirow{2}{*}{\textbf{Atom Chirality Tag}} & Unspecified, Tetrahedral cw, \\
&  Tetrahedral ccw, Other \\
\midrule
\textbf{Bond Type} & Single, Double, Triple, Aromatic\\
\midrule
\textbf{Bond Direction} & -, Endupright, Enddownright\\
\bottomrule
\end{tabular}}}}
\label{tab:features}
\end{table}

In this section, we conduct extensive experiments on two public datasets (Tox21 and Sider) to compare performances of different models and show related analysis.



\begin{table*}[th]
\centering
\caption{
The performances of all methods on both datasets. 
Our proposed method Meta-MGNN can outperform all baseline methods. The last column reports the average improvements (in percentage) of Meta-MGNN over the best baseline method in different tasks. Bond indicates the best performance. Underline represents the best baseline performance.
}
\vspace{-0.1in}
\setlength{\tabcolsep}{1.2mm}{
{\scalebox{1.0}{%
\begin{tabular}{l|c||cccccc||cc}
\toprule
{\multirow{2}{*}{\textbf{Dataset}}} & {\multirow{2}{*}{\textbf{Task}}}  & \bf{GraphSAGE ~\cite{hamilton2017inductive}} & \bf{GCN ~\cite{kipf2017semi}} & \bf{MAML~\cite{finn2017model} } &  \bf{Seq3seq ~\cite{xu2019powerful}} &  \bf{EGNN~\cite{kim2019edge}} & \bf{PreGNN ~\cite{hu2020strategies}} &  {\multirow{2}{*}{\textbf{Meta-MGNN}}} & {\multirow{2}{*}{\textbf{$\Delta$AUC}}}\\
&& \bf{(2017)} & \bf{(2017)} & \bf{(2017)} &  \bf{(2018)} &  \bf{(2019)} & \bf{(2020)} &\\ 
\midrule
\rowcolor{gray!20} & {} & \multicolumn{8}{c}{\textbf{1-shot}}  \\
\midrule
{\multirow{4}{*}{\textbf{Tox21}}} & \textbf{SR-HS} &65.97 &65.00 &68.56 &\underline{73.18} &72.51 &73.09 &\bf{73.81} & \textbf{+0.63} \\ 
& \textbf{SR-MMP} &71.23 &71.20 &76.34 &\underline{79.08} &76.90 &76.20 & \bf{79.09} & \textbf{+0.01} \\ 
& \textbf{SR-p53} &58.05 &66.60 &71.28 &75.23 &\underline{\textbf{78.03}} &76.87 &77.71 &-0.32 \\ 
& \textbf{Average} &65.10 &67.60 &72.06 &\underline{75.83} &75.81 &75.39 &\bf{76.87} & \textbf{+1.04} \\
\midrule
{\multirow{7}{*}{\textbf{Sider}}} & \textbf{Si-T1} &65.23 &63.60 &66.82 &66.50 &71.39 &\underline{73.04} &\bf{75.41} & \textbf{+2.37}\\ 
& \textbf{Si-T2} &60.47 &62.01 &63.62 &57.03 &\underline{67.87} &66.06 &\bf{69.39} & \textbf{+1.52} \\ 
& \textbf{Si-T3} &61.45 &64.52 &67.50 &61.38 &68.23 &\underline{70.36} &\bf{70.65} & \textbf{+0.29} \\ 
& \textbf{Si-T4} &64.41 &65.28 &69.02 &63.45 &\underline{72.67} &72.34 & \textbf{72.69} & \textbf{+0.02} \\ 
& \textbf{Si-T5} &77.85 &74.95 &77.07 &74.83 &\underline{78.88} &77.99 &\bf{79.95} & \textbf{+1.07} \\ 
& \textbf{Si-T6} &61.19 &63.20 &67.01 &63.70 &66.31 &\underline{69.45} &\bf{71.97} & \textbf{+2.52} \\ 
& \textbf{Average} &65.10 &65.60 &68.51 &64.48 &70.89 &\underline{71.54} &\bf{73.34} & \textbf{+1.80} \\
\midrule
\rowcolor{gray!20}
& {} & \multicolumn{8}{c}{\textbf{5-shots}} \\
\midrule
{\multirow{4}{*}{\textbf{Tox21}}} & \textbf{SR-HS} &69.09 &68.13 &69.02 &\underline{74.07} &73.23 &73.39 &\bf{74.80} & \textbf{+0.73} \\ 
& \textbf{SR-MMP} &72.22 &69.06 &76.43 &\underline{\textbf{80.40}} &79.07 &78.25 &80.26 & -0.14  \\
& \textbf{SR-p53} &61.45 &72.01 &73.95 &77.07 &\underline{78.12} &78.01 &\bf{79.00} & \textbf{+0.88}  \\
& \textbf{Average} &67.59 &69.73 &73.13 &\underline{77.18} &76.81 &76.55 &\bf{78.02} & \textbf{+0.84}  \\
\midrule
{\multirow{7}{*}{\textbf{Sider}}} &
\textbf{Si-T1} &67.61 &65.66 &70.12 &68.99 &72.76 &\underline{74.77} &\bf{76.32} & \textbf{+1.55}  \\ 
& \textbf{Si-T2} &59.86 &64.62 &64.46 &56.53 &\underline{68.13} &65.69 &\bf{69.34} & \textbf{+1.21}  \\
& \textbf{Si-T3} &60.61 &64.90 &68.20 &64.20 &70.11 &\underline{71.07} &\bf{72.29} & \textbf{+1.22}  \\
& \textbf{Si-T4} &64.82 &64.85 &67.75 &67.15 &72.73 &\underline{73.42} &\bf{74.46} & \textbf{+1.04}  \\
& \textbf{Si-T5} &78.33 &76.93 &78.61 &78.55 &79.61 &\underline{80.67} &\bf{81.79} & \textbf{+1.12} \\
& \textbf{Si-T6} &61.91 &62.06 &67.74 &66.30 &67.17 &\underline{71.48} &\bf{74.12} & \textbf{+2.64}  \\
& \textbf{Average} &65.52 &66.50 &69.48 &66.95 &71.75 &\underline{72.85} &\bf{74.72} & \textbf{+1.87}  \\
\bottomrule
\end{tabular}}}}
\label{tab:baseline}
\end{table*}

\subsection{Datasets}
We evaluate different methods on the Tox21 and Sider datasets. They are collected from MoleculeNet~\cite{wu2018moleculenet}, which is a large scale benchmark dataset for molecular machine learning. Tox21 has 7,831 instances with 12 different tasks; Sider has 1,427 instances with 27 different tasks.
In each task, molecules are divided into positive instances and negative instances (i.e., binary labels). A positive instance means that a molecule has a specific property, and a negative instance means that a molecule does not have the property. We manually split 3 tasks from Tox21 and 6 tasks from Sider for meta-testing. The details of these two datasets are as follows:

\begin{itemize}[leftmargin=*]
    \item \textbf{Tox21}\footnote{https://tripod.nih.gov/tox21/
challenge/}: Toxicity on 12 biological targets, including nuclear receptors and stress response pathways.
    \item \textbf{Sider}: 27 system organ classes where molecules are marketed drugs and adverse drug reactions~\cite{kuhn2016sider}.
\end{itemize}

\noindent\textbf{Dataset Processing.} The raw data of molecules are given as \textit{SMILES} strings. We transfer \textit{SMILES} strings to molecular graphs by using Rdkit.Chem~\cite{landrum2013rdkit}. Then we extracted a set of node and bond features which can preserve the molecular structure best to use in the experiments. The details about features are listed in Table~\ref{tab:features}.

\subsection{Baselines}
We compare our model with multiple baseline models.

\begin{itemize}[leftmargin=*]
    \item \textbf{GraphSAGE~\cite{hamilton2017inductive}.} It generates the nodes' embedding by sampling and aggregating their neighbors' embeddings, which can effectively capture the graph information. 
    \item \textbf{GCN~\cite{kipf2017semi}.} It is a widely used graph-based model, which contains an effective convolutional neural network component. GCN outperforms various models by learning both local graph structure and features of nodes.
    \item \textbf{MAML~\cite{finn2017model}.} It builds a task-agnostic algorithm for few-shot learning, where training a model's parameters using a small number of gradient updates will lead to fast learning on new tasks.
    \item \textbf{Seq3seq~\cite{zhang2018seq3seq}.}  It is a Seq2Seq model for molecular property prediction. The loss function contains both self-recovery loss and inference task loss. 
    \item \textbf{EGNN~\cite{kim2019edge}. } It is an edge-labeling graph neural network for few-shot learning which is proved a well-generalizable model for low-data problem.
    \item \textbf{PreGNN~\cite{hu2020strategies}.} This model develops self-supervised learning to pretrain GNN for molecular property prediction. It captures both useful local and global information. 
\end{itemize}

\begin{figure*}[th]
\centering
\vspace{-0.1in}
\subfigure[SR-HS]
{\includegraphics[width=0.32\textwidth,height=3.4cm]{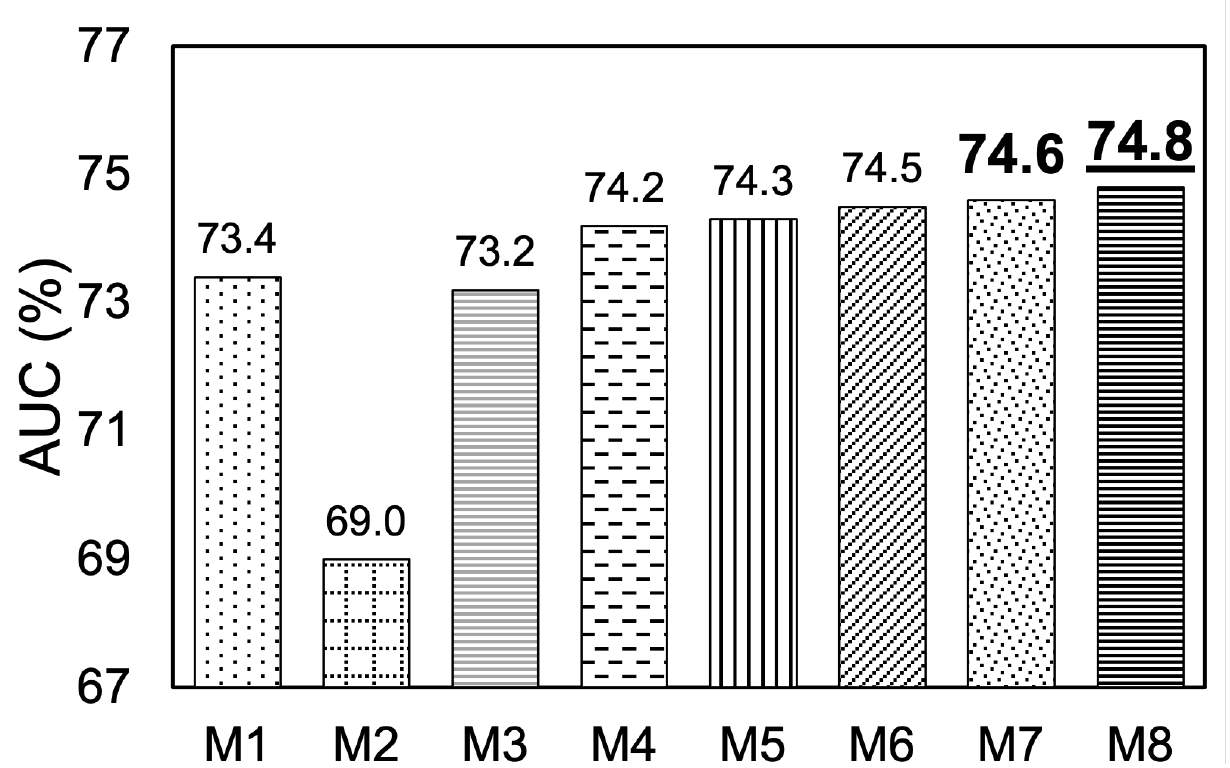}\label{fig:template3}}
\subfigure[SR-MMP]
{\includegraphics[width=0.32\textwidth,height=3.4cm]{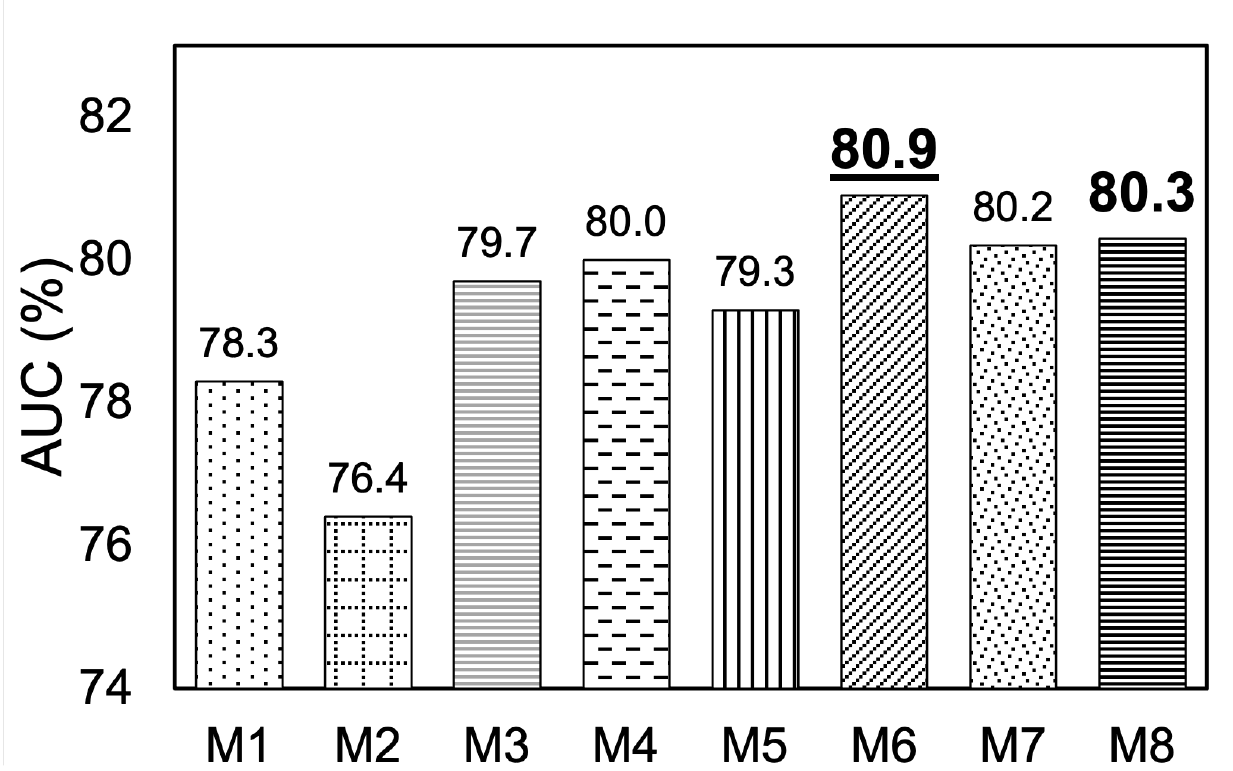}\label{fig:template4}}
\subfigure[SR-p53]
{\includegraphics[width=0.32\textwidth,height=3.4cm]{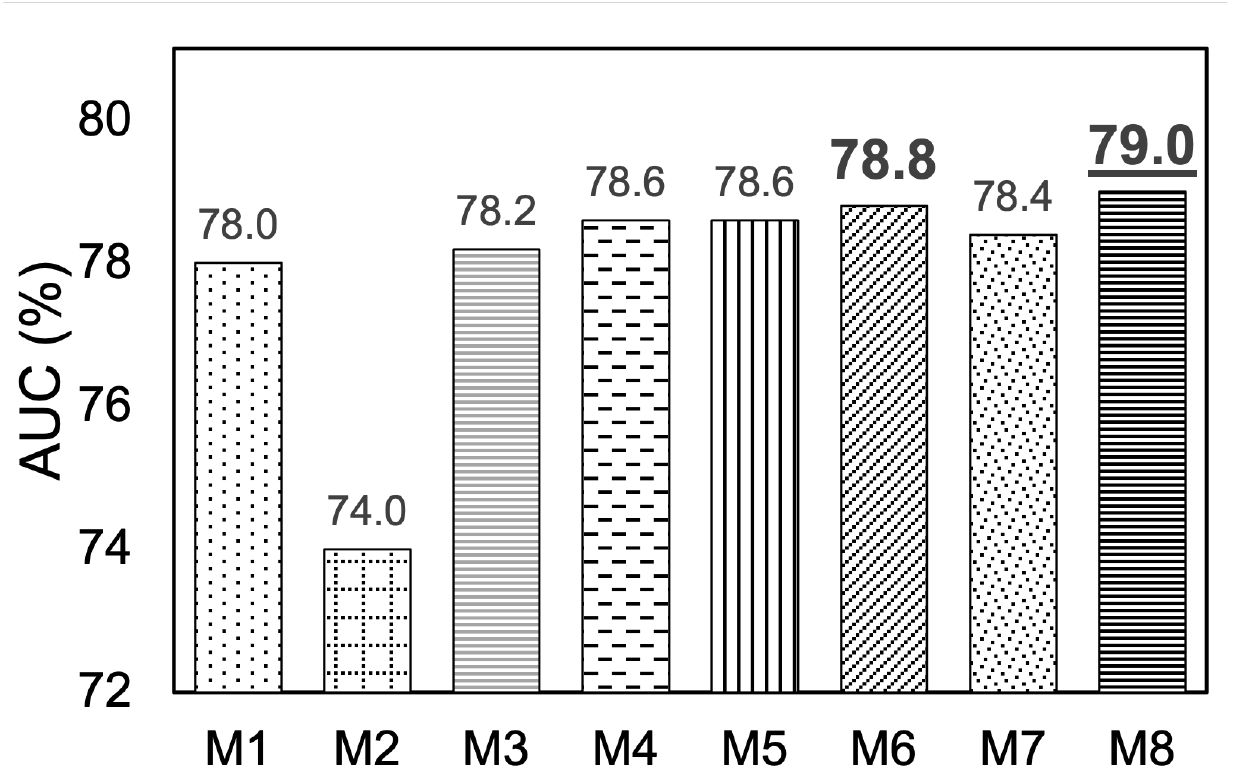}\label{fig:template1}}
\subfigure[Si-T1]
{\includegraphics[width=0.32\textwidth,height=3.4cm]{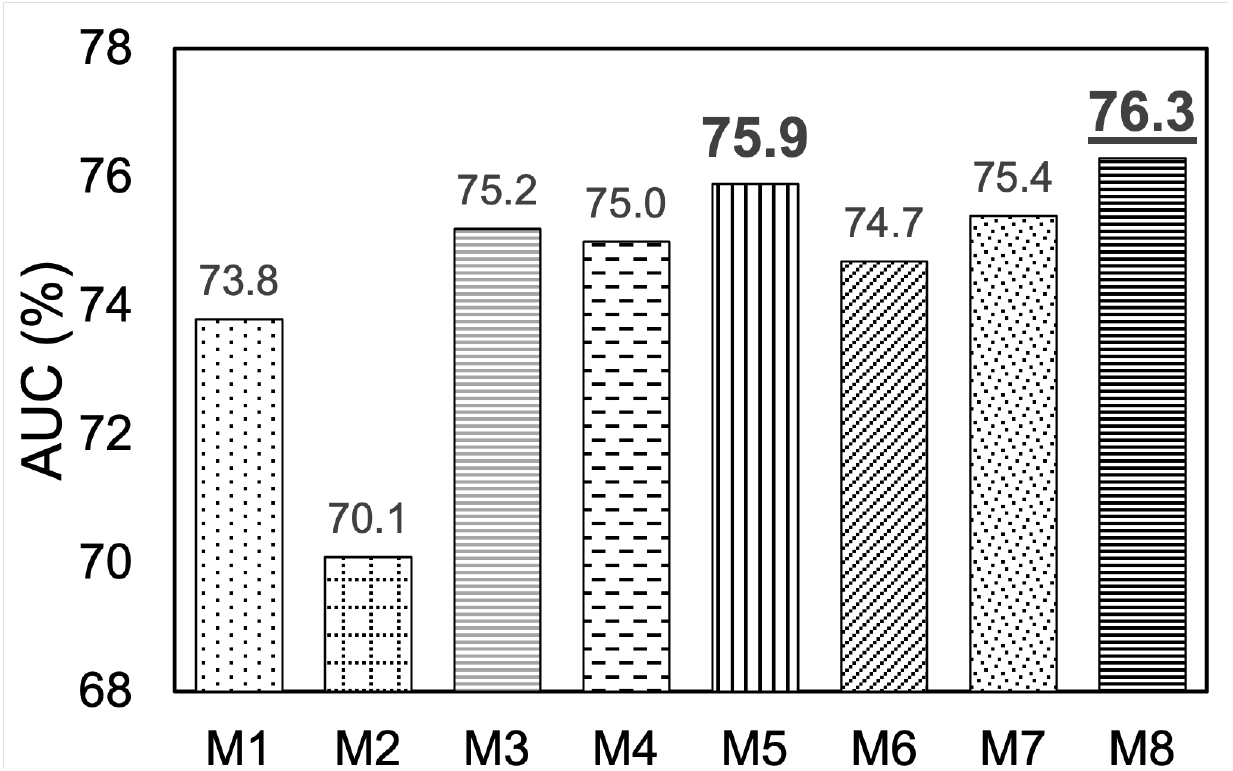}\label{fig:template3}}
\subfigure[Si-T2]
{\includegraphics[width=0.32\textwidth,height=3.4cm]{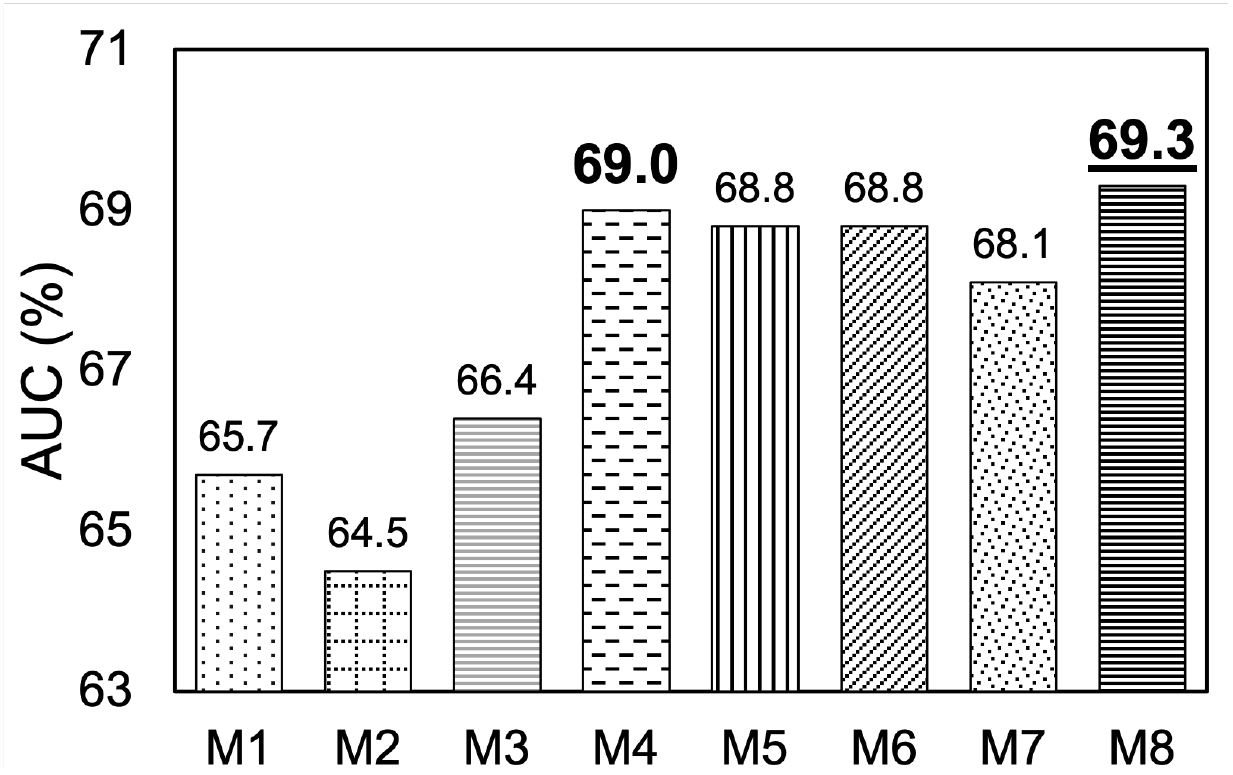}\label{fig:template4}}
\subfigure[Si-T3]
{\includegraphics[width=0.32\textwidth,height=3.4cm]{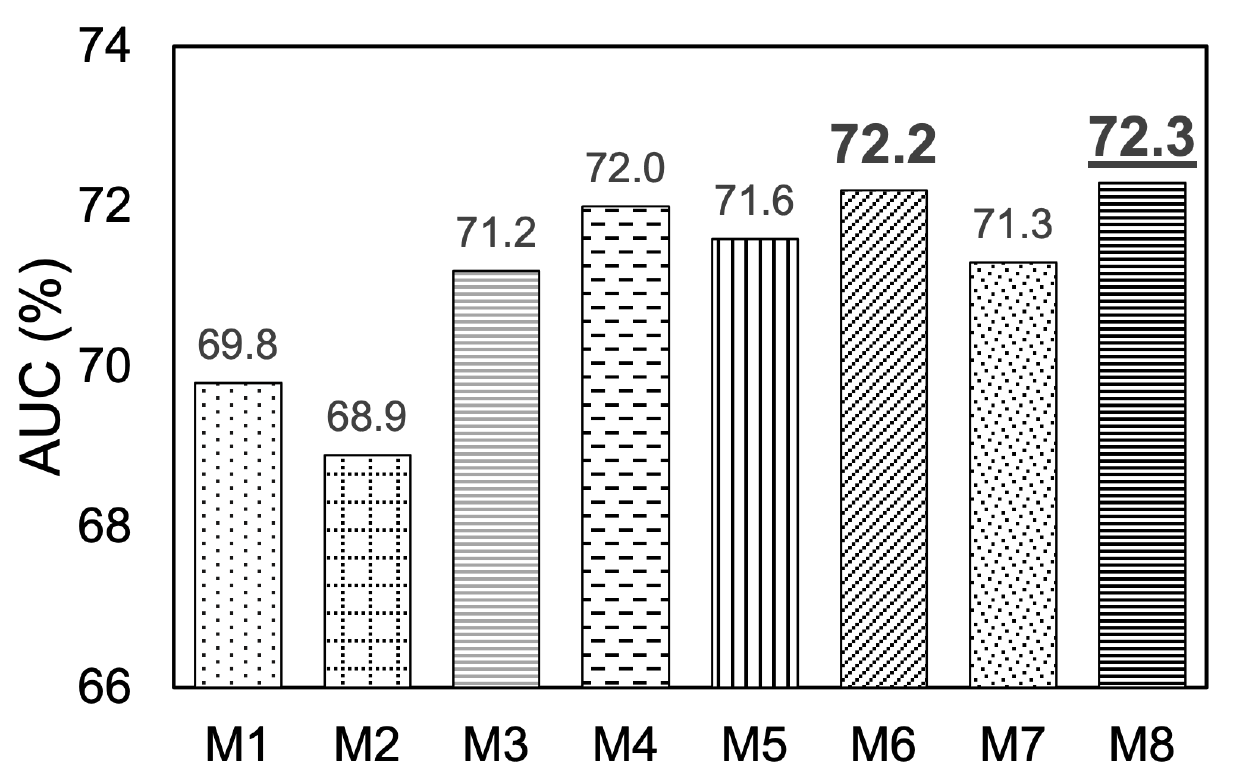}\label{fig:template1}}
\subfigure[Si-T4]
{\includegraphics[width=0.32\textwidth,height=3.4cm]{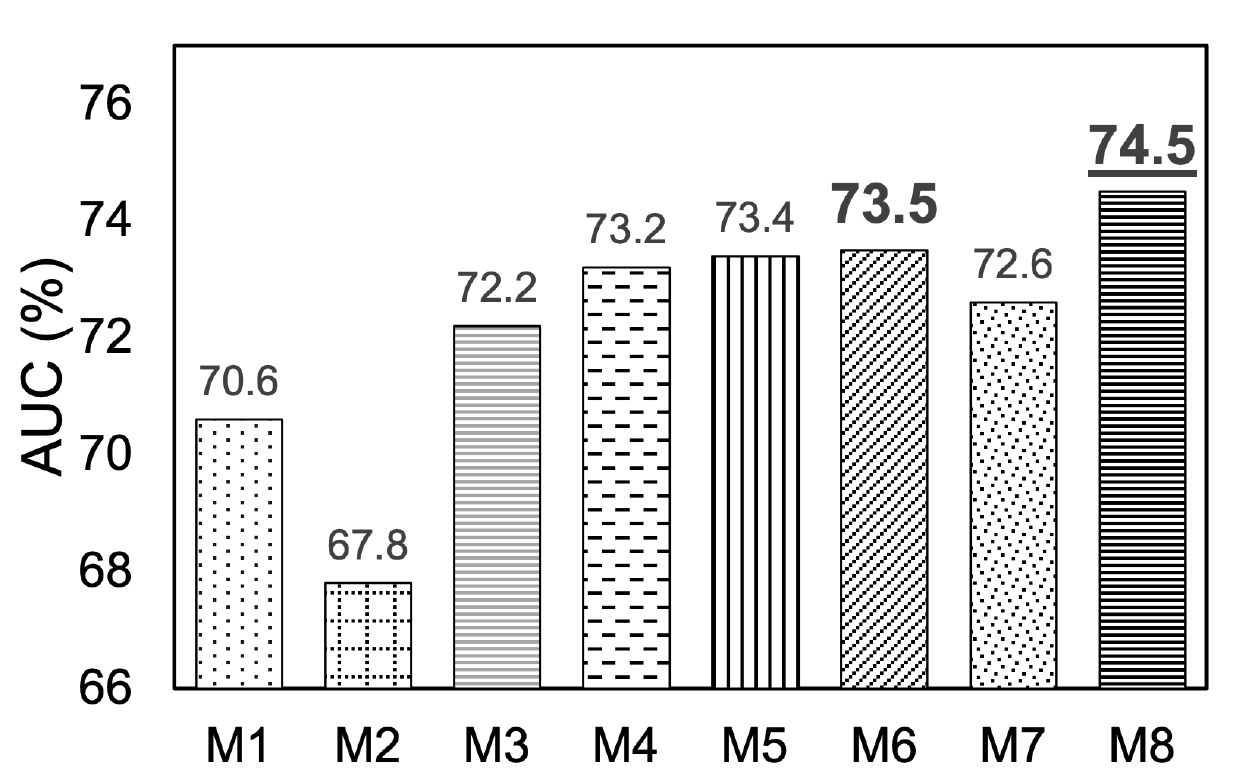}\label{fig:template1}}
\subfigure[Si-T5]
{\includegraphics[width=0.32\textwidth,height=3.4cm]{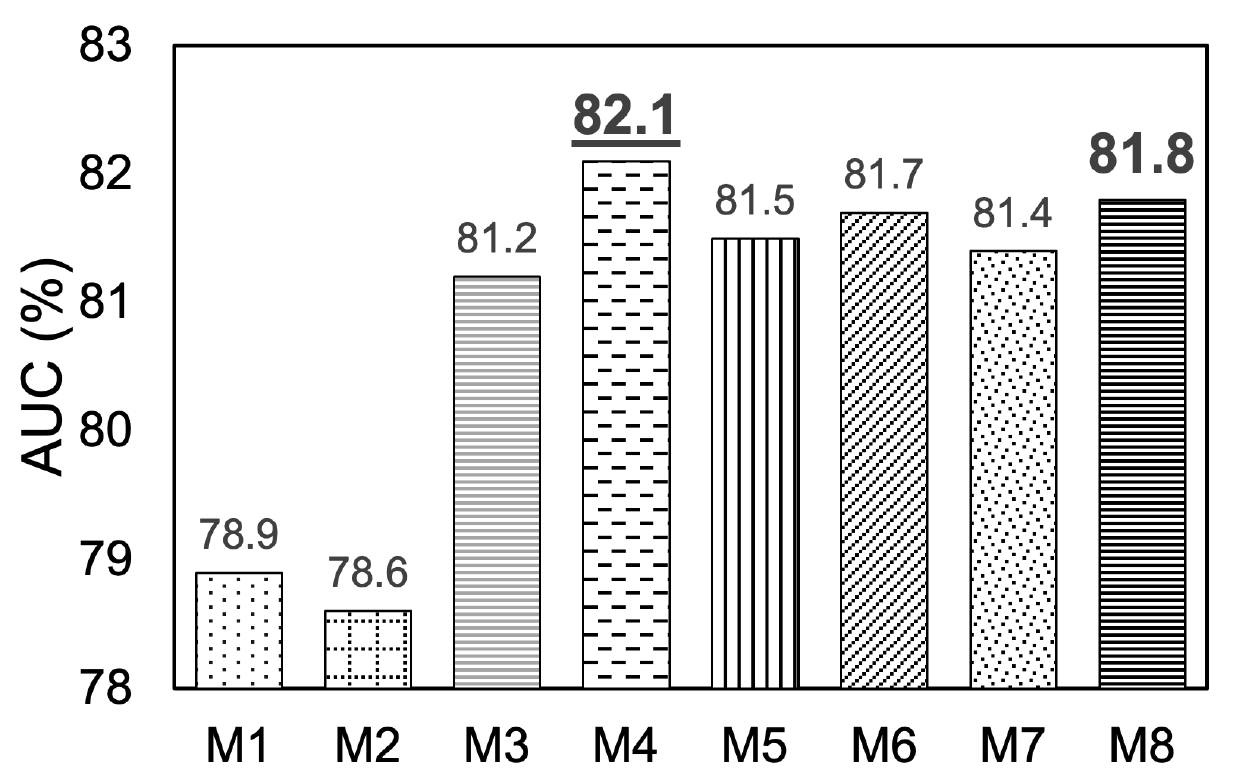}\label{fig:template3}}
\subfigure[Si-T6]
{\includegraphics[width=0.32\textwidth,height=3.4cm]{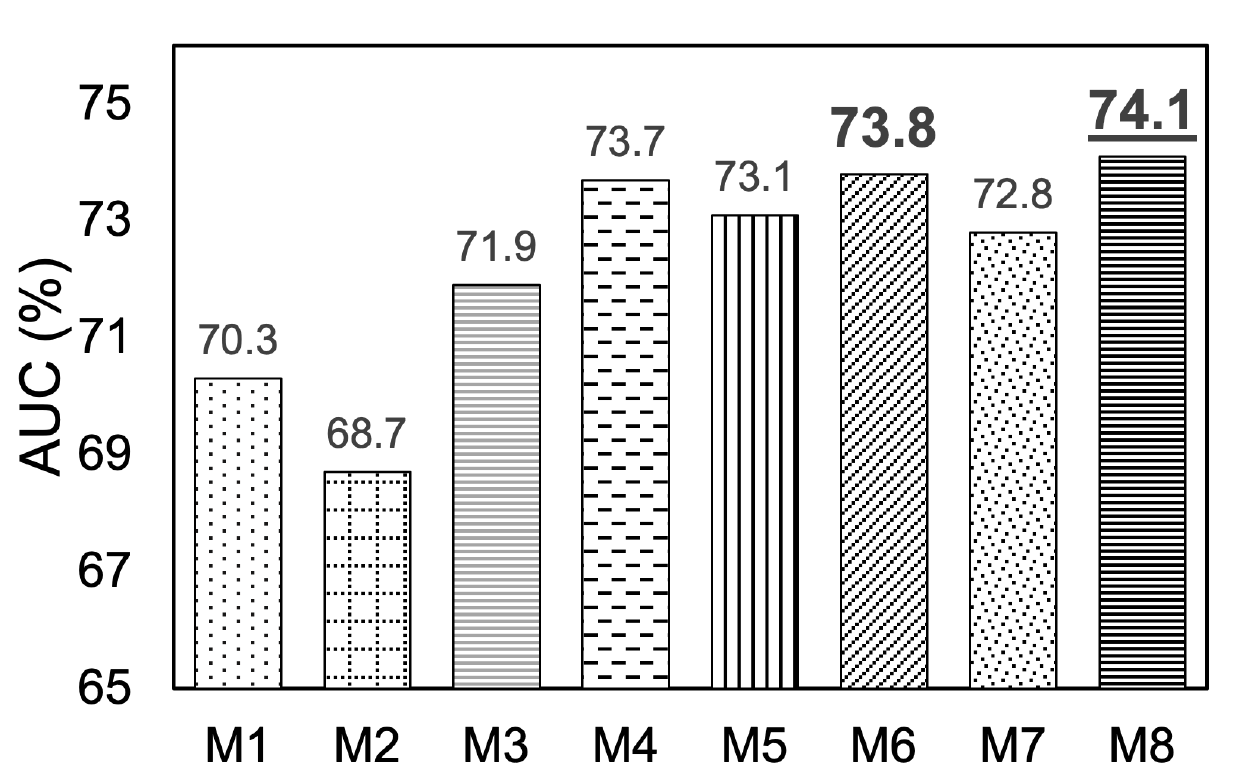}\label{fig:template4}}
\vspace{-0.15in}
\caption{Performances of different model variants in Tox21 data (the first row) and Sider (the remaining figures) data. Different model components (i.e., graph neural network pre-training, self-supervised module, and task-aware attention) indeed make effect and improve the model performance. Our proposed model (M8) has better result than all other model variants.}
\label{fig:ablation_result}
\end{figure*}

\subsection{Evaluation Metrics} We evaluate the performance of each model using ROC-AUC. We consider each molecular property as an independent task for few-shot learning. We use 3 and 6 tasks as test tasks of Tox21 and Sider data, respectively. Each task is a binary label classification task. Table \ref{tab:baseline} and Figure \ref{tab:Ablation_study} report the result for each test task. For both datasets, we consider 2-way classification with 1 and 5 shots. 

\subsection{Reproducibility Settings} 
We take graph isomorphism network (GIN)~\cite{xu2019powerful} as base graph neural network. In our experiment, we  utilize the supervised-contextpred pre-trained GIN of PreGNN~\cite{hu2020strategies}.
The GIN layer number is set as 5. We set all embedding dimensions to 300. The same feature will share the same initial embedding. We set the update step in training tasks as 5 and the update step in testing tasks as 10. We set the trade-off weight of self-supervised module as 0.1. We use Pytorch to implement the model and run it on a GPU. 

\subsection{Comparisons with Baselines}
\vspace{0.05in}
\textbf{Overall Performance.} The overall performances of all methods are reported  in Table~\ref{tab:baseline}. According to this table, we can find that Meta-MGNN outperforms all baseline models on both Tox21 and Sider datasets. Specifically, for 1-shot learning, the average improvements are \textbf{+1.04\%} and \textbf{+1.80\%}  on Tox21 and on Sider, respectively. The values equal \textbf{+0.84\%} and \textbf{+1.87\%} for 5-shot learning. 
In addition, we observe that PreGNN~\cite{hu2020strategies} and EGNN~\cite{kim2019edge} perform the best among all baseline methods on average. 
However, the baseline methods do not have stable performance on different tasks. In other words, they may perform well on one task, but perform poorly on another task. In comparison, the performance of Meta-MGNN is stable. It has the best performance for all tasks in both datasest.

\vspace{0.05in}
\noindent{\textbf{Analyzing Meta-MGNN Structure. }} 
MAML demonstrates superior performance than the other two GNN models (GraphSage and GCN). It makes sense since MAML trains the model through meta-learning, making it better adapt to 
new tasks with few data samples. Besides taking advantage of meta-learning,
Meta-MGNN also utilizes pre-trained graph neural network model (PreGNN)~\cite{hu2020strategies} to initialize model parameters. PreGNN uses a large amount of molecules data to pre-train the graph neural network, which leads to better parameter initialization. 
Therefore, by taking advantages of both meta-learning and pre-training, Meta-MGNN demonstrates superior performance than the other baseline methods. 


\begin{table}[t]
\centering
\caption{We implemented 8 model variants for ablation study. The abbreviations used in the table: pre-trained model (PTM), meta-learning (ML), bond reconstruction (BR), atom-type prediction (AP) and task-aware attention (T-At). }
\vspace{-0.1in}
\setlength{\tabcolsep}{3.3mm}{
\begin{tabular}{l||ccccc}
\toprule
{} &PTM &ML &BR &AP &T-At\\ 
\midrule 
\textbf{M1}  &\cmarkb & & & & \\
\textbf{M2}  & &\cmarkb & & &  \\
\textbf{M3}  &\cmarkb &\cmarkb & & &  \\ 
\textbf{M4}  &\cmarkb &\cmarkb &\cmarkb & &  \\
\textbf{M5}  &\cmarkb &\cmarkb & &\cmarkb &  \\
\textbf{M6}  &\cmarkb &\cmarkb &\cmarkb &\cmarkb &  \\
\textbf{M7}  &\cmarkb &\cmarkb & & &\cmarkb  \\
\textbf{M8}  &\cmarkb &\cmarkb &\cmarkb &\cmarkb &\cmarkb  \\
\bottomrule
\end{tabular}}
\vspace{-0.1in}
\label{tab:Ablation_study}
\end{table}

\begin{figure*}[t]
    \centering
    \subfigure[Meta-MGNN]
    {\includegraphics[width=0.33\textwidth]{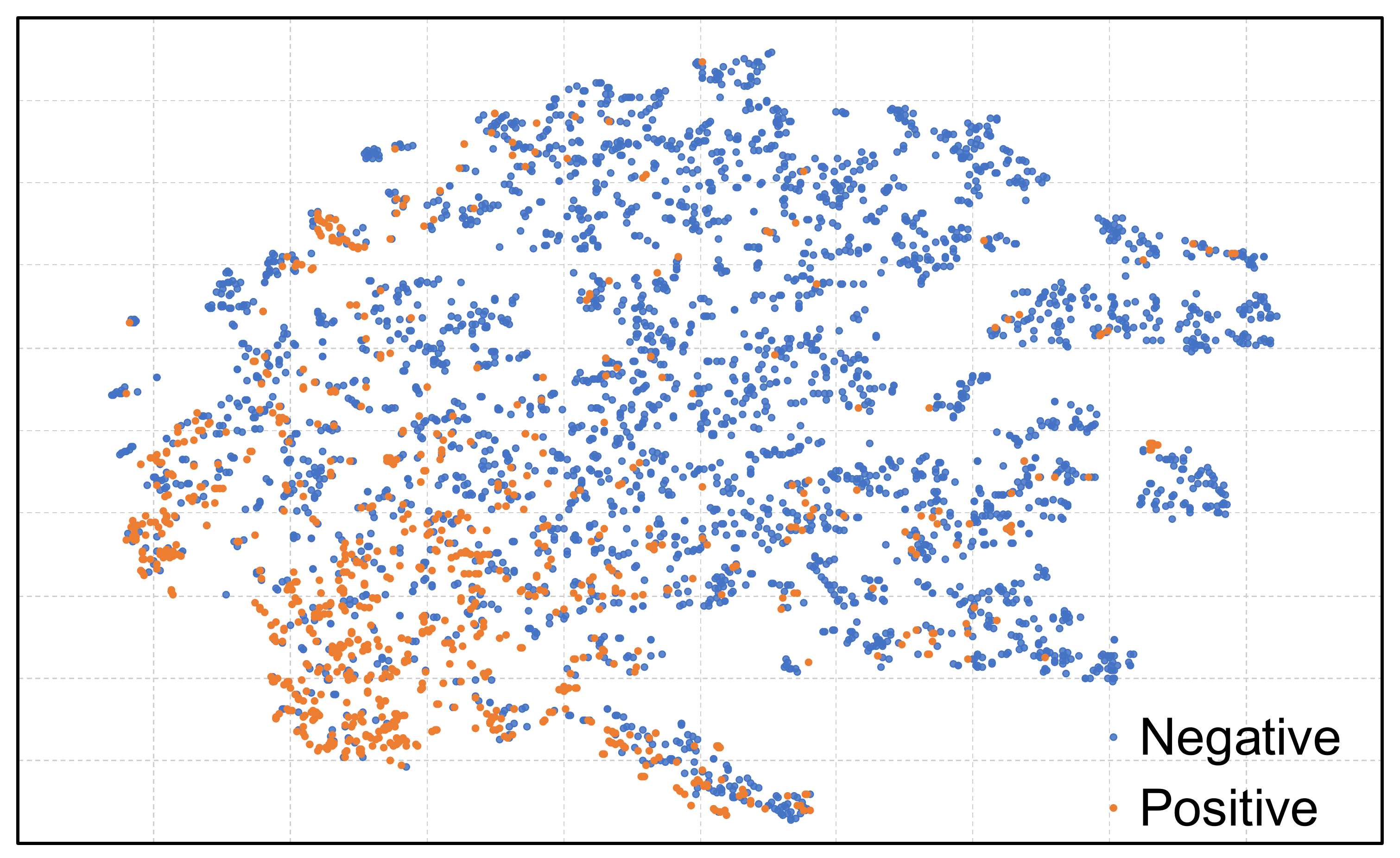}\label{fig:template3}}
    \subfigure[PreGNN~\cite{hu2020strategies}]
    {\includegraphics[ width=0.33\textwidth]{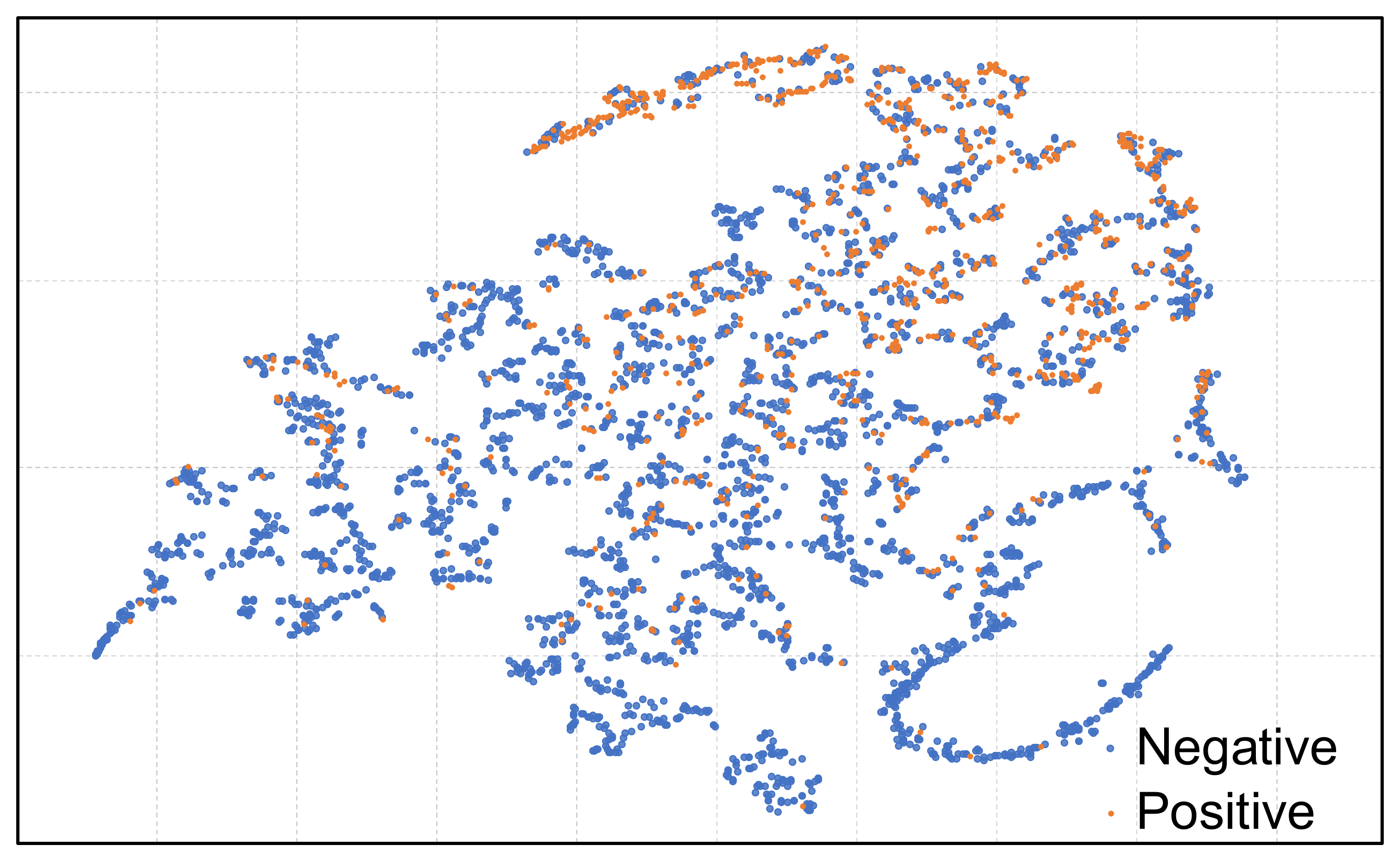}\label{fig:template4}}
    \subfigure[MAML~\cite{finn2017model}]
    {\includegraphics[ width=0.33\textwidth]{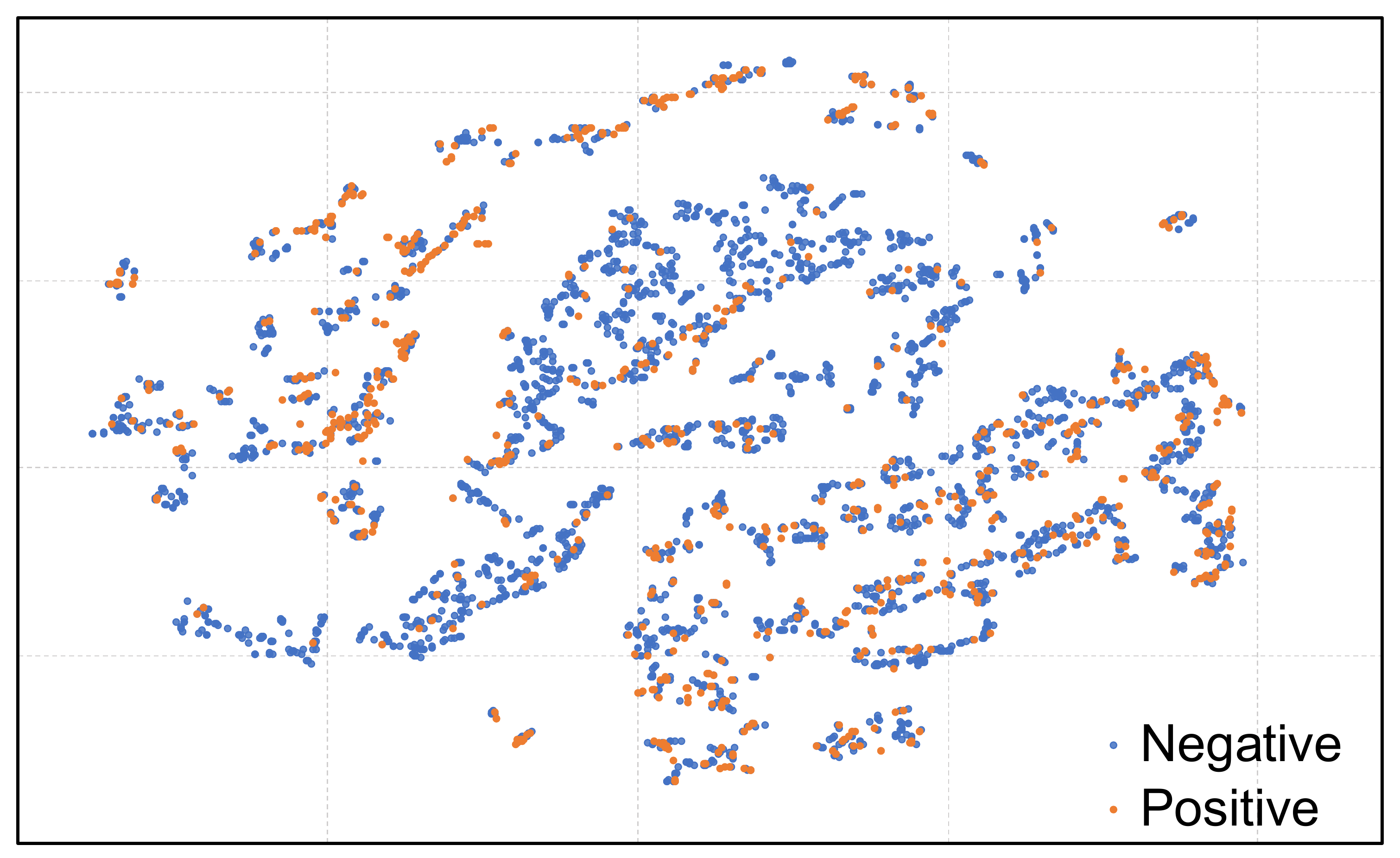}\label{fig:template4}}
    \vspace{-0.15in}
    \caption{Visualizations of molecular embeddings generated by our model (Meta-MGNN), PreGNN~\cite{hu2020strategies}, and MAML~\cite{finn2017model}. The blue dots denote negative labels in SR-MMP (a molecular property). The orange dots represent positive labels in SR-MMP. Our model can better discriminate embeddings of these two kinds of labels than the other methods.}
    \label{fig:embed}
    \vspace{-0.15in}
\end{figure*}

\noindent\textbf{Performance on Different Datasets. }
From Table~\ref{tab:baseline}, we can observe that our proposed Meta-MGNN outperforms the best baseline method by \textbf{+1.80\%} for 1-shot learning and \textbf{+1.87\%} for 5-shots learning on Sider dataset. However, the average improvements on Tox21 are \textbf{+1.04\%} for 1-shot learning and \textbf{+0.84\%} for 5-shots learning, which are smaller than those in Sider. As we know, the major advantage of few-shot learning model is to make model predict new tasks better by using a small number of data samples. Obviously, training with more tasks allows the model to learn more knowledge. The advantages of the few-shot learning model can be better reflected on the dataset which contains more tasks. Since Sider has more tasks than Tox21, Meta-MGNN can deliver greater improvements on Sider than on Tox21. 
Additionally, we also find that the overall performance on Tox21 is better than that on Sider for all models. This is due to the larger size of Tox21, which improves the generalization capabilities of these deep learning models, as reflected by the evaluation scores.

\subsection{\textbf{Ablation Study}} 
\textbf{Settings.} Besides comparing with baseline methods, we also implement model variants (ablation studies) to show the effectiveness of different model components. The details of different model variants are illustrated as follows (also shown in the Table~\ref{tab:Ablation_study}): 
\begin{itemize}[leftmargin=*]
    \item \textbf{M1.} Pre-trained graph neural network. We take GIN~\cite{xu2019powerful} as our base graph neural network and pre-train it by both supervised and unsupervised (Context Prediction) pre-training strategies.
    \item \textbf{M2.} Graph neural network model (without pre-training) trained with the meta learning process. 
    \item \textbf{M3.} Our base model, which is based on GIN~\cite{xu2019powerful} and learned with meta-learning algorithms. This model is also pre-trained by supervised and unsupervised (Context Prediction) pre-training.
    \item \textbf{M4 \& M5 \& M6.} These models are based on M3 and augmented with the self-supervised module. M4, M5, and M6 are augmented with the bond reconstruction, the atom-type prediction, and both of them, respectively. This is to analyze the effectiveness of the self-supervised module.
    \item \textbf{M7.} It is based on M3 and enhanced with task-aware attention to incorporate the importance of different tasks. This is to analyze the effectiveness of task weight in meta-learning. 
    \item \textbf{M8.} It is based on M3 and augmented with both both self-supervised module and self-attentive task weight. 
\end{itemize}

\noindent\textbf{Performance Comparison and Analysis. } 
The performances of all model variants are shown in Figure~\ref{fig:ablation_result}. The three sub-figures in the first row are model performance on Tox21. The sub-figures in the second row and third row are model performance on Sider. 
There are several findings from these figures. First, M2 has the worst results in all cases, illustrating the significant impact of the pre-training step for graph neural networks. Second, the performance of M3 is better than M1 and M2, which indicates the effectiveness of combining both the pre-training and few-shot learning strategies. Third, adding different self-supervised components (bond reconstruction and atom type prediction) can further improve model performance, as reflected by the better performances of M4, M5, and M6 over M3. Among these three variants, M6 has the best performance as it adds both self-supervised tasks. Additionally, M7 outperforms M3, demonstrating the benefit of incorporating task-attention weight into the meta-learning process. At last, M8 (the proposed model) has the best performances in most cases, which shows the best capability graph neural network model trained by meta-learning process and augmented with both self-supervised module and task-aware attention. 
According to these findings, we can conclude that different model components indeed bring benefits to model design and improve performance.

\subsection{Case Study of Embedding Visualization}
To better show the effectiveness of our model, we visualize the molecular embeddings generated by our proposed Meta-MGNN, PreGNN~\cite{hu2020strategies} and MAML~\cite{finn2017model} using t-SNE~\cite{maaten2008visualizing}, which are shown in Figure~\ref{fig:embed}. Specifically, it shows the embedding result of testing datasets from SR-MMP (a molecular property). The blue plots and orange plots represent molecules without SR-MMP property and with SR-MMP property, respectively. It can be observed that our model achieves better performance in discriminating two kinds of molecules than the other two models. In Figure~\ref{fig:embed}(a), the bottom left corner of the figure is mostly occupied by orange dots and the blues ones are mostly in the upper right corner. However, most orange plots are mixed with blue plots in Figure~\ref{fig:embed}(b) and ~\ref{fig:embed}(c).

\section{Conclusions}
In this work, we proposed a few-shot learning approach for the molecular property prediction problem, which is important and has not been well studied. We proposed a novel model called Meta-MGNN. Meta-MGNN utilized a graph neural network (with pre-training) to learn molecular embeddings and further employed a meta-learning process to learn well-initialized model parameters that could be fast adapted to new molecular properties with few-shot data samples. A self-supervised module and self-attentive task weight were further proposed and incorporate into the meta-learning framework, which benefited the whole model. We evaluated our model on two public multi-task datasets and the comparison of the experimental results showed that our model can outperform state-of-art methods. The effectiveness of each model component was also verified. The initial success of this study suggests following studies. The future work might consider better task embedding formulation when computing task weight in meta-learning. It is also possible to fuse both graph and sequence model to learn molecular embeddings for the  meta-learning process. 

\section*{Acknowledgements}
This work was supported in part by National Science Foundation grant CCI-1925607. 
We thank the anonymous referees for their valuable comments and helpful suggestions. 

\balance
\bibliographystyle{ACM-Reference-Format}
\bibliography{reference}


\begin{thebibliography}{46}


\ifx \showCODEN    \undefined \def \showCODEN     #1{\unskip}     \fi
\ifx \showDOI      \undefined \def \showDOI       #1{#1}\fi
\ifx \showISBNx    \undefined \def \showISBNx     #1{\unskip}     \fi
\ifx \showISBNxiii \undefined \def \showISBNxiii  #1{\unskip}     \fi
\ifx \showISSN     \undefined \def \showISSN      #1{\unskip}     \fi
\ifx \showLCCN     \undefined \def \showLCCN      #1{\unskip}     \fi
\ifx \shownote     \undefined \def \shownote      #1{#1}          \fi
\ifx \showarticletitle \undefined \def \showarticletitle #1{#1}   \fi
\ifx \showURL      \undefined \def \showURL       {\relax}        \fi
\providecommand\bibfield[2]{#2}
\providecommand\bibinfo[2]{#2}
\providecommand\natexlab[1]{#1}
\providecommand\showeprint[2][]{arXiv:#2}

\bibitem[\protect\citeauthoryear{Altae-Tran, Ramsundar, Pappu, and
  Pande}{Altae-Tran et~al\mbox{.}}{2017}]%
        {altae2017low}
\bibfield{author}{\bibinfo{person}{Han Altae-Tran}, \bibinfo{person}{Bharath
  Ramsundar}, \bibinfo{person}{Aneesh~S Pappu}, {and} \bibinfo{person}{Vijay
  Pande}.} \bibinfo{year}{2017}\natexlab{}.
\newblock \showarticletitle{Low data drug discovery with one-shot learning}. In
  \bibinfo{booktitle}{\emph{ACS Central Science}}.
\newblock


\bibitem[\protect\citeauthoryear{Bertinetto, Henriques, Torr, and
  Vedaldi}{Bertinetto et~al\mbox{.}}{2018}]%
        {bertinetto2018meta}
\bibfield{author}{\bibinfo{person}{Luca Bertinetto}, \bibinfo{person}{Joao~F
  Henriques}, \bibinfo{person}{Philip~HS Torr}, {and} \bibinfo{person}{Andrea
  Vedaldi}.} \bibinfo{year}{2018}\natexlab{}.
\newblock \showarticletitle{Meta-learning with differentiable closed-form
  solvers}. In \bibinfo{booktitle}{\emph{International Conference for Learning
  Representation (ICLR)}}.
\newblock


\bibitem[\protect\citeauthoryear{Devlin, Chang, Lee, and Toutanova}{Devlin
  et~al\mbox{.}}{2018}]%
        {devlin2018bert}
\bibfield{author}{\bibinfo{person}{Jacob Devlin}, \bibinfo{person}{Ming-Wei
  Chang}, \bibinfo{person}{Kenton Lee}, {and} \bibinfo{person}{Kristina
  Toutanova}.} \bibinfo{year}{2018}\natexlab{}.
\newblock \showarticletitle{Bert: Pre-training of deep bidirectional
  transformers for language understanding}. In
  \bibinfo{booktitle}{\emph{Conference of the North American Chapter of the
  Association for Computational Linguistics: Human Language Technologies
  (NAACL-HLT)}}.
\newblock


\bibitem[\protect\citeauthoryear{Fan, Ma, Li, He, Zhao, Tang, and Yin}{Fan
  et~al\mbox{.}}{2019}]%
        {fan2019graph}
\bibfield{author}{\bibinfo{person}{Wenqi Fan}, \bibinfo{person}{Yao Ma},
  \bibinfo{person}{Qing Li}, \bibinfo{person}{Yuan He}, \bibinfo{person}{Eric
  Zhao}, \bibinfo{person}{Jiliang Tang}, {and} \bibinfo{person}{Dawei Yin}.}
  \bibinfo{year}{2019}\natexlab{}.
\newblock \showarticletitle{Graph neural networks for social recommendation}.
  In \bibinfo{booktitle}{\emph{The Web Conference}}.
\newblock


\bibitem[\protect\citeauthoryear{Finn, Abbeel, and Levine}{Finn
  et~al\mbox{.}}{2017}]%
        {finn2017model}
\bibfield{author}{\bibinfo{person}{Chelsea Finn}, \bibinfo{person}{Pieter
  Abbeel}, {and} \bibinfo{person}{Sergey Levine}.}
  \bibinfo{year}{2017}\natexlab{}.
\newblock \showarticletitle{Model-agnostic meta-learning for fast adaptation of
  deep networks}. In \bibinfo{booktitle}{\emph{International Conference on
  Machine Learning (ICML)}}.
\newblock


\bibitem[\protect\citeauthoryear{Garcia and Bruna}{Garcia and Bruna}{2018}]%
        {garcia2018few}
\bibfield{author}{\bibinfo{person}{Victor Garcia} {and} \bibinfo{person}{Joan
  Bruna}.} \bibinfo{year}{2018}\natexlab{}.
\newblock \showarticletitle{Few-shot learning with graph neural networks}. In
  \bibinfo{booktitle}{\emph{International Conference for Learning
  Representation (ICLR)}}.
\newblock


\bibitem[\protect\citeauthoryear{Gilmer, Schoenholz, Riley, Vinyals, and
  Dahl}{Gilmer et~al\mbox{.}}{2017}]%
        {gilmer2017neural}
\bibfield{author}{\bibinfo{person}{Justin Gilmer}, \bibinfo{person}{Samuel~S
  Schoenholz}, \bibinfo{person}{Patrick~F Riley}, \bibinfo{person}{Oriol
  Vinyals}, {and} \bibinfo{person}{George~E Dahl}.}
  \bibinfo{year}{2017}\natexlab{}.
\newblock \showarticletitle{Neural message passing for quantum chemistry}. In
  \bibinfo{booktitle}{\emph{International Conference on Machine Learning
  (ICML)}}.
\newblock


\bibitem[\protect\citeauthoryear{Guo, Yu, Zhang, Jiang, and Chawla}{Guo
  et~al\mbox{.}}{2020}]%
        {guo2020graseq}
\bibfield{author}{\bibinfo{person}{Zhichun Guo}, \bibinfo{person}{Wenhao Yu},
  \bibinfo{person}{Chuxu Zhang}, \bibinfo{person}{Meng Jiang}, {and}
  \bibinfo{person}{Nitesh Chawla}.} \bibinfo{year}{2020}\natexlab{}.
\newblock \showarticletitle{{GraSeq}: graph and sequence fusion learning for
  molecular property prediction}. In \bibinfo{booktitle}{\emph{International
  Conference on Information and Knowledge Management (CIKM)}}.
\newblock


\bibitem[\protect\citeauthoryear{Hamilton, Ying, and Leskovec}{Hamilton
  et~al\mbox{.}}{2017}]%
        {hamilton2017inductive}
\bibfield{author}{\bibinfo{person}{Will Hamilton}, \bibinfo{person}{Zhitao
  Ying}, {and} \bibinfo{person}{Jure Leskovec}.}
  \bibinfo{year}{2017}\natexlab{}.
\newblock \showarticletitle{Inductive representation learning on large graphs}.
  In \bibinfo{booktitle}{\emph{Advances in Neural Information Processing
  Systems}}.
\newblock


\bibitem[\protect\citeauthoryear{Hu, Liu, Gomes, Zitnik, Liang, Pande, and
  Leskovec}{Hu et~al\mbox{.}}{2020}]%
        {hu2020strategies}
\bibfield{author}{\bibinfo{person}{Weihua Hu}, \bibinfo{person}{Bowen Liu},
  \bibinfo{person}{Joseph Gomes}, \bibinfo{person}{Marinka Zitnik},
  \bibinfo{person}{Percy Liang}, \bibinfo{person}{Vijay Pande}, {and}
  \bibinfo{person}{Jure Leskovec}.} \bibinfo{year}{2020}\natexlab{}.
\newblock \showarticletitle{Strategies for pre-training graph neural networks}.
  In \bibinfo{booktitle}{\emph{International Conference for Learning
  Representation (ICLR)}}.
\newblock


\bibitem[\protect\citeauthoryear{Kim, Kim, Kim, and Yoo}{Kim
  et~al\mbox{.}}{2019}]%
        {kim2019edge}
\bibfield{author}{\bibinfo{person}{Jongmin Kim}, \bibinfo{person}{Taesup Kim},
  \bibinfo{person}{Sungwoong Kim}, {and} \bibinfo{person}{Chang~D Yoo}.}
  \bibinfo{year}{2019}\natexlab{}.
\newblock \showarticletitle{Edge-labeling graph neural network for few-shot
  learning}. In \bibinfo{booktitle}{\emph{IEEE Conference on Computer Vision
  and Pattern Recognition (CVPR)}}.
\newblock


\bibitem[\protect\citeauthoryear{Kipf and Welling}{Kipf and Welling}{2017}]%
        {kipf2017semi}
\bibfield{author}{\bibinfo{person}{Thomas~N Kipf} {and} \bibinfo{person}{Max
  Welling}.} \bibinfo{year}{2017}\natexlab{}.
\newblock \showarticletitle{Semi-supervised classification with graph
  convolutional networks}. In \bibinfo{booktitle}{\emph{International
  Conference for Learning Representation}}.
\newblock


\bibitem[\protect\citeauthoryear{Kuhn, Letunic, Jensen, and Bork}{Kuhn
  et~al\mbox{.}}{2016}]%
        {kuhn2016sider}
\bibfield{author}{\bibinfo{person}{Michael Kuhn}, \bibinfo{person}{Ivica
  Letunic}, \bibinfo{person}{Lars~Juhl Jensen}, {and} \bibinfo{person}{Peer
  Bork}.} \bibinfo{year}{2016}\natexlab{}.
\newblock \showarticletitle{The SIDER database of drugs and side effects}. In
  \bibinfo{booktitle}{\emph{Nucleic Acids Research}}.
\newblock


\bibitem[\protect\citeauthoryear{Landrum}{Landrum}{2013}]%
        {landrum2013rdkit}
\bibfield{author}{\bibinfo{person}{Greg Landrum}.}
  \bibinfo{year}{2013}\natexlab{}.
\newblock \bibinfo{title}{Rdkit: A software suite for cheminformatics,
  computational chemistry, and predictive modeling}.
\newblock
\newblock


\bibitem[\protect\citeauthoryear{Lee and Choi}{Lee and Choi}{2018}]%
        {lee2018gradient}
\bibfield{author}{\bibinfo{person}{Yoonho Lee} {and} \bibinfo{person}{Seungjin
  Choi}.} \bibinfo{year}{2018}\natexlab{}.
\newblock \showarticletitle{Gradient-based meta-learning with learned layerwise
  metric and subspace}. In \bibinfo{booktitle}{\emph{International Conference
  on Machine Learning}}.
\newblock


\bibitem[\protect\citeauthoryear{Lin, Feng, Santos, Yu, Xiang, Zhou, and
  Bengio}{Lin et~al\mbox{.}}{2017}]%
        {lin2017structured}
\bibfield{author}{\bibinfo{person}{Zhouhan Lin}, \bibinfo{person}{Minwei Feng},
  \bibinfo{person}{Cicero Nogueira~dos Santos}, \bibinfo{person}{Mo Yu},
  \bibinfo{person}{Bing Xiang}, \bibinfo{person}{Bowen Zhou}, {and}
  \bibinfo{person}{Yoshua Bengio}.} \bibinfo{year}{2017}\natexlab{}.
\newblock \showarticletitle{A structured self-attentive sentence embedding}. In
  \bibinfo{booktitle}{\emph{International Conference for Learning
  Representation (ICLR)}}.
\newblock


\bibitem[\protect\citeauthoryear{Liu, Sun, Jia, Ma, Xing, Wu, Gao, Sun,
  Boulnois, and Fan}{Liu et~al\mbox{.}}{2019}]%
        {liu2019chemi}
\bibfield{author}{\bibinfo{person}{Ke Liu}, \bibinfo{person}{Xiangyan Sun},
  \bibinfo{person}{Lei Jia}, \bibinfo{person}{Jun Ma}, \bibinfo{person}{Haoming
  Xing}, \bibinfo{person}{Junqiu Wu}, \bibinfo{person}{Hua Gao},
  \bibinfo{person}{Yax Sun}, \bibinfo{person}{Florian Boulnois}, {and}
  \bibinfo{person}{Jie Fan}.} \bibinfo{year}{2019}\natexlab{}.
\newblock \showarticletitle{Chemi-Net: a molecular graph convolutional network
  for accurate drug property prediction}. In
  \bibinfo{booktitle}{\emph{International Journal of Molecular Sciences}}.
\newblock


\bibitem[\protect\citeauthoryear{Lu, Liu, Wang, Huang, Lin, and He}{Lu
  et~al\mbox{.}}{2019}]%
        {lu2019molecular}
\bibfield{author}{\bibinfo{person}{Chengqiang Lu}, \bibinfo{person}{Qi Liu},
  \bibinfo{person}{Chao Wang}, \bibinfo{person}{Zhenya Huang},
  \bibinfo{person}{Peize Lin}, {and} \bibinfo{person}{Lixin He}.}
  \bibinfo{year}{2019}\natexlab{}.
\newblock \showarticletitle{Molecular property prediction: A multilevel quantum
  interactions modeling perspective}. In \bibinfo{booktitle}{\emph{AAAI
  Conference on Artificial Intelligence (AAAI)}}.
\newblock


\bibitem[\protect\citeauthoryear{Maaten and Hinton}{Maaten and Hinton}{2008}]%
        {maaten2008visualizing}
\bibfield{author}{\bibinfo{person}{Laurens van~der Maaten} {and}
  \bibinfo{person}{Geoffrey Hinton}.} \bibinfo{year}{2008}\natexlab{}.
\newblock \showarticletitle{Visualizing data using t-SNE}. In
  \bibinfo{booktitle}{\emph{Journal of Machine Learning Research}}.
\newblock


\bibitem[\protect\citeauthoryear{Mansimov, Mahmood, Kang, and Cho}{Mansimov
  et~al\mbox{.}}{2019}]%
        {mansimov2019molecular}
\bibfield{author}{\bibinfo{person}{Elman Mansimov}, \bibinfo{person}{Omar
  Mahmood}, \bibinfo{person}{Seokho Kang}, {and} \bibinfo{person}{Kyunghyun
  Cho}.} \bibinfo{year}{2019}\natexlab{}.
\newblock \showarticletitle{Molecular geometry prediction using a deep
  generative graph neural network}. In \bibinfo{booktitle}{\emph{Scientific
  Reports}}.
\newblock


\bibitem[\protect\citeauthoryear{Mikolov, Sutskever, Chen, Corrado, and
  Dean}{Mikolov et~al\mbox{.}}{2013}]%
        {mikolov2013distributed}
\bibfield{author}{\bibinfo{person}{Tomas Mikolov}, \bibinfo{person}{Ilya
  Sutskever}, \bibinfo{person}{Kai Chen}, \bibinfo{person}{Greg~S Corrado},
  {and} \bibinfo{person}{Jeff Dean}.} \bibinfo{year}{2013}\natexlab{}.
\newblock \showarticletitle{Distributed representations of words and phrases
  and their compositionality}. In \bibinfo{booktitle}{\emph{Advances in Neural
  Information Processing Systems (NeurIPS)}}.
\newblock


\bibitem[\protect\citeauthoryear{Riniker and Landrum}{Riniker and
  Landrum}{2013}]%
        {riniker2013similarity}
\bibfield{author}{\bibinfo{person}{Sereina Riniker} {and}
  \bibinfo{person}{Gregory~A Landrum}.} \bibinfo{year}{2013}\natexlab{}.
\newblock \showarticletitle{Similarity maps-a visualization strategy for
  molecular fingerprints and machine-learning methods}. In
  \bibinfo{booktitle}{\emph{Journal of Cheminformatics}}.
\newblock


\bibitem[\protect\citeauthoryear{Sliwoski, Kothiwale, Meiler, and
  Lowe}{Sliwoski et~al\mbox{.}}{2014}]%
        {sliwoski2014computational}
\bibfield{author}{\bibinfo{person}{Gregory Sliwoski},
  \bibinfo{person}{Sandeepkumar Kothiwale}, \bibinfo{person}{Jens Meiler},
  {and} \bibinfo{person}{Edward~W Lowe}.} \bibinfo{year}{2014}\natexlab{}.
\newblock \showarticletitle{Computational methods in drug discovery}. In
  \bibinfo{booktitle}{\emph{Pharmacological Reviews}}.
\newblock


\bibitem[\protect\citeauthoryear{Song, Xiao, Wang, Charlin, Zhang, and
  Tang}{Song et~al\mbox{.}}{2019}]%
        {song2019session}
\bibfield{author}{\bibinfo{person}{Weiping Song}, \bibinfo{person}{Zhiping
  Xiao}, \bibinfo{person}{Yifan Wang}, \bibinfo{person}{Laurent Charlin},
  \bibinfo{person}{Ming Zhang}, {and} \bibinfo{person}{Jian Tang}.}
  \bibinfo{year}{2019}\natexlab{}.
\newblock \showarticletitle{Session-based social recommendation via dynamic
  graph attention networks}. In \bibinfo{booktitle}{\emph{The ACM International
  Conference on Web Search and Data Mining}}.
\newblock


\bibitem[\protect\citeauthoryear{Sung, Yang, Zhang, Xiang, Torr, and
  Hospedales}{Sung et~al\mbox{.}}{2018}]%
        {sung2018learning}
\bibfield{author}{\bibinfo{person}{Flood Sung}, \bibinfo{person}{Yongxin Yang},
  \bibinfo{person}{Li Zhang}, \bibinfo{person}{Tao Xiang},
  \bibinfo{person}{Philip~HS Torr}, {and} \bibinfo{person}{Timothy~M
  Hospedales}.} \bibinfo{year}{2018}\natexlab{}.
\newblock \showarticletitle{Learning to compare: relation network for few-shot
  learning}. In \bibinfo{booktitle}{\emph{IEEE Conference on Computer Vision
  and Pattern Recognition (CVPR)}}.
\newblock


\bibitem[\protect\citeauthoryear{Vamathevan, Clark, Czodrowski, Dunham, Ferran,
  Lee, Li, Madabhushi, Shah, Spitzer, et~al\mbox{.}}{Vamathevan
  et~al\mbox{.}}{2019}]%
        {vamathevan2019applications}
\bibfield{author}{\bibinfo{person}{Jessica Vamathevan},
  \bibinfo{person}{Dominic Clark}, \bibinfo{person}{Paul Czodrowski},
  \bibinfo{person}{Ian Dunham}, \bibinfo{person}{Edgardo Ferran},
  \bibinfo{person}{George Lee}, \bibinfo{person}{Bin Li},
  \bibinfo{person}{Anant Madabhushi}, \bibinfo{person}{Parantu Shah},
  \bibinfo{person}{Michaela Spitzer}, {et~al\mbox{.}}}
  \bibinfo{year}{2019}\natexlab{}.
\newblock \showarticletitle{Applications of machine learning in drug discovery
  and development}. In \bibinfo{booktitle}{\emph{Nature Reviews Drug
  Discovery}}.
\newblock


\bibitem[\protect\citeauthoryear{Veli{\v{c}}kovi{\'c}, Cucurull, Casanova,
  Romero, Lio, and Bengio}{Veli{\v{c}}kovi{\'c} et~al\mbox{.}}{2018}]%
        {velivckovic2018graph}
\bibfield{author}{\bibinfo{person}{Petar Veli{\v{c}}kovi{\'c}},
  \bibinfo{person}{Guillem Cucurull}, \bibinfo{person}{Arantxa Casanova},
  \bibinfo{person}{Adriana Romero}, \bibinfo{person}{Pietro Lio}, {and}
  \bibinfo{person}{Yoshua Bengio}.} \bibinfo{year}{2018}\natexlab{}.
\newblock \showarticletitle{Graph attention networks}. In
  \bibinfo{booktitle}{\emph{International Conference for Learning
  Representation (ICLR)}}.
\newblock


\bibitem[\protect\citeauthoryear{Vinyals, Blundell, Lillicrap, Wierstra,
  et~al\mbox{.}}{Vinyals et~al\mbox{.}}{2016}]%
        {vinyals2016matching}
\bibfield{author}{\bibinfo{person}{Oriol Vinyals}, \bibinfo{person}{Charles
  Blundell}, \bibinfo{person}{Timothy Lillicrap}, \bibinfo{person}{Daan
  Wierstra}, {et~al\mbox{.}}} \bibinfo{year}{2016}\natexlab{}.
\newblock \showarticletitle{Matching networks for one shot learning}. In
  \bibinfo{booktitle}{\emph{Advances in Neural Information Processing Systems
  (NeurIPS)}}.
\newblock


\bibitem[\protect\citeauthoryear{Wang, Guo, Wang, Sun, and Huang}{Wang
  et~al\mbox{.}}{2019}]%
        {wang2019smiles}
\bibfield{author}{\bibinfo{person}{Sheng Wang}, \bibinfo{person}{Yuzhi Guo},
  \bibinfo{person}{Yuhong Wang}, \bibinfo{person}{Hongmao Sun}, {and}
  \bibinfo{person}{Junzhou Huang}.} \bibinfo{year}{2019}\natexlab{}.
\newblock \showarticletitle{SMILES-BERT: Large scale unsupervised pre-training
  for molecular property prediction}. In \bibinfo{booktitle}{\emph{ACM
  International Conference on Bioinformatics, Computational Biology, and Health
  Informatics (BCB)}}.
\newblock


\bibitem[\protect\citeauthoryear{Waring, Arrowsmith, Leach, Leeson, Mandrell,
  Owen, Pairaudeau, Pennie, Pickett, Wang, et~al\mbox{.}}{Waring
  et~al\mbox{.}}{2015}]%
        {waring2015analysis}
\bibfield{author}{\bibinfo{person}{Michael~J Waring}, \bibinfo{person}{John
  Arrowsmith}, \bibinfo{person}{Andrew~R Leach}, \bibinfo{person}{Paul~D
  Leeson}, \bibinfo{person}{Sam Mandrell}, \bibinfo{person}{Robert~M Owen},
  \bibinfo{person}{Garry Pairaudeau}, \bibinfo{person}{William~D Pennie},
  \bibinfo{person}{Stephen~D Pickett}, \bibinfo{person}{Jibo Wang},
  {et~al\mbox{.}}} \bibinfo{year}{2015}\natexlab{}.
\newblock \showarticletitle{An analysis of the attrition of drug candidates
  from four major pharmaceutical companies}. In
  \bibinfo{booktitle}{\emph{Nature Reviews Drug Discovery}}.
\newblock


\bibitem[\protect\citeauthoryear{Weininger, Weininger, and Weininger}{Weininger
  et~al\mbox{.}}{1989}]%
        {weininger1989smiles}
\bibfield{author}{\bibinfo{person}{David Weininger}, \bibinfo{person}{Arthur
  Weininger}, {and} \bibinfo{person}{Joseph~L Weininger}.}
  \bibinfo{year}{1989}\natexlab{}.
\newblock \showarticletitle{SMILES. 2. Algorithm for generation of unique
  SMILES notation}. In \bibinfo{booktitle}{\emph{Journal of Chemical
  Information and Computer Sciences}}.
\newblock


\bibitem[\protect\citeauthoryear{Wu, Pan, Chen, Long, Zhang, and Philip}{Wu
  et~al\mbox{.}}{2020}]%
        {wu2020comprehensive}
\bibfield{author}{\bibinfo{person}{Zonghan Wu}, \bibinfo{person}{Shirui Pan},
  \bibinfo{person}{Fengwen Chen}, \bibinfo{person}{Guodong Long},
  \bibinfo{person}{Chengqi Zhang}, {and} \bibinfo{person}{S~Yu Philip}.}
  \bibinfo{year}{2020}\natexlab{}.
\newblock \showarticletitle{A comprehensive survey on graph neural networks}.
  In \bibinfo{booktitle}{\emph{IEEE Transactions on Neural Networks and
  Learning Systems}}.
\newblock


\bibitem[\protect\citeauthoryear{Wu, Ramsundar, Feinberg, Gomes, Geniesse,
  Pappu, Leswing, and Pande}{Wu et~al\mbox{.}}{2018}]%
        {wu2018moleculenet}
\bibfield{author}{\bibinfo{person}{Zhenqin Wu}, \bibinfo{person}{Bharath
  Ramsundar}, \bibinfo{person}{Evan~N Feinberg}, \bibinfo{person}{Joseph
  Gomes}, \bibinfo{person}{Caleb Geniesse}, \bibinfo{person}{Aneesh~S Pappu},
  \bibinfo{person}{Karl Leswing}, {and} \bibinfo{person}{Vijay Pande}.}
  \bibinfo{year}{2018}\natexlab{}.
\newblock \showarticletitle{{MoleculeNet}: a benchmark for molecular machine
  learning}. In \bibinfo{booktitle}{\emph{Chemical Science}}.
\newblock


\bibitem[\protect\citeauthoryear{Xu, Hu, Leskovec, and Jegelka}{Xu
  et~al\mbox{.}}{2019}]%
        {xu2019powerful}
\bibfield{author}{\bibinfo{person}{Keyulu Xu}, \bibinfo{person}{Weihua Hu},
  \bibinfo{person}{Jure Leskovec}, {and} \bibinfo{person}{Stefanie Jegelka}.}
  \bibinfo{year}{2019}\natexlab{}.
\newblock \showarticletitle{How powerful are graph neural networks?}. In
  \bibinfo{booktitle}{\emph{International Conference for Learning
  Representation (ICLR)}}.
\newblock


\bibitem[\protect\citeauthoryear{Xu, Wang, Zhu, and Huang}{Xu
  et~al\mbox{.}}{2017}]%
        {xu2017seq2seq}
\bibfield{author}{\bibinfo{person}{Zheng Xu}, \bibinfo{person}{Sheng Wang},
  \bibinfo{person}{Feiyun Zhu}, {and} \bibinfo{person}{Junzhou Huang}.}
  \bibinfo{year}{2017}\natexlab{}.
\newblock \showarticletitle{Seq2seq fingerprint: An unsupervised deep molecular
  embedding for drug discovery}. In \bibinfo{booktitle}{\emph{ACM International
  Conference on Bioinformatics, Computational Biology, and Health Informatics
  (BCB)}}.
\newblock


\bibitem[\protect\citeauthoryear{Yao, Liu, Wei, Tang, and Li}{Yao
  et~al\mbox{.}}{2019}]%
        {yao2019learning}
\bibfield{author}{\bibinfo{person}{Huaxiu Yao}, \bibinfo{person}{Yiding Liu},
  \bibinfo{person}{Ying Wei}, \bibinfo{person}{Xianfeng Tang}, {and}
  \bibinfo{person}{Zhenhui Li}.} \bibinfo{year}{2019}\natexlab{}.
\newblock \showarticletitle{Learning from multiple cities: A meta-learning
  approach for spatial-temporal prediction}. In \bibinfo{booktitle}{\emph{The
  Web Conference (WWW)}}.
\newblock


\bibitem[\protect\citeauthoryear{Yao, Zhang, Wei, Jiang, Wang, Huang, Chawla,
  and Li}{Yao et~al\mbox{.}}{2020}]%
        {yao2020graph}
\bibfield{author}{\bibinfo{person}{Huaxiu Yao}, \bibinfo{person}{Chuxu Zhang},
  \bibinfo{person}{Ying Wei}, \bibinfo{person}{Meng Jiang},
  \bibinfo{person}{Suhang Wang}, \bibinfo{person}{Junzhou Huang},
  \bibinfo{person}{Nitesh Chawla}, {and} \bibinfo{person}{Zhenhui Li}.}
  \bibinfo{year}{2020}\natexlab{}.
\newblock \showarticletitle{Graph few-shot learning via knowledge transfer}. In
  \bibinfo{booktitle}{\emph{AAAI Conference on Artificial Intelligence
  (AAAI)}}.
\newblock


\bibitem[\protect\citeauthoryear{Yu, Yu, Zhao, and Jiang}{Yu
  et~al\mbox{.}}{2020}]%
        {yu2020identifying}
\bibfield{author}{\bibinfo{person}{Wenhao Yu}, \bibinfo{person}{Mengxia Yu},
  \bibinfo{person}{Tong Zhao}, {and} \bibinfo{person}{Meng Jiang}.}
  \bibinfo{year}{2020}\natexlab{}.
\newblock \showarticletitle{Identifying referential intention with
  heterogeneous contexts}. In \bibinfo{booktitle}{\emph{The Web Conference
  (WWW)}}.
\newblock


\bibitem[\protect\citeauthoryear{Zhang, Song, Huang, Swami, and Chawla}{Zhang
  et~al\mbox{.}}{2019}]%
        {zhang2019heterogeneous}
\bibfield{author}{\bibinfo{person}{Chuxu Zhang}, \bibinfo{person}{Dongjin
  Song}, \bibinfo{person}{Chao Huang}, \bibinfo{person}{Ananthram Swami}, {and}
  \bibinfo{person}{Nitesh~V Chawla}.} \bibinfo{year}{2019}\natexlab{}.
\newblock \showarticletitle{Heterogeneous graph neural network}. In
  \bibinfo{booktitle}{\emph{ACM SIGKDD International Conference on Knowledge
  Discovery and Data Mining (KDD)}}.
\newblock


\bibitem[\protect\citeauthoryear{Zhang, Yao, Huang, Jiang, Li, and
  Chawla}{Zhang et~al\mbox{.}}{2020a}]%
        {zhang2020few}
\bibfield{author}{\bibinfo{person}{Chuxu Zhang}, \bibinfo{person}{Huaxiu Yao},
  \bibinfo{person}{Chao Huang}, \bibinfo{person}{Meng Jiang},
  \bibinfo{person}{Zhenhui Li}, {and} \bibinfo{person}{Nitesh~V Chawla}.}
  \bibinfo{year}{2020}\natexlab{a}.
\newblock \showarticletitle{Few-shot knowledge graph completion}. In
  \bibinfo{booktitle}{\emph{AAAI Conference on Artificial Intelligence
  (AAAI)}}.
\newblock


\bibitem[\protect\citeauthoryear{Zhang, Yu, Saebi, Jiang, and Chawla}{Zhang
  et~al\mbox{.}}{2020b}]%
        {zhang2020few2}
\bibfield{author}{\bibinfo{person}{Chuxu Zhang}, \bibinfo{person}{Lu Yu},
  \bibinfo{person}{Mandana Saebi}, \bibinfo{person}{Meng Jiang}, {and}
  \bibinfo{person}{Nitesh Chawla}.} \bibinfo{year}{2020}\natexlab{b}.
\newblock \showarticletitle{Few-shot multi-hop relation reasoning over
  knowledge bases}. In \bibinfo{booktitle}{\emph{Conference on Empirical
  Methods in Natural Language Processing: Findings (EMNLP: Findings)}}.
\newblock


\bibitem[\protect\citeauthoryear{Zhang, Che, Ghahramani, Bengio, and
  Song}{Zhang et~al\mbox{.}}{2018a}]%
        {zhang2018metagan}
\bibfield{author}{\bibinfo{person}{Ruixiang Zhang}, \bibinfo{person}{Tong Che},
  \bibinfo{person}{Zoubin Ghahramani}, \bibinfo{person}{Yoshua Bengio}, {and}
  \bibinfo{person}{Yangqiu Song}.} \bibinfo{year}{2018}\natexlab{a}.
\newblock \showarticletitle{Metagan: An adversarial approach to few-shot
  learning}. In \bibinfo{booktitle}{\emph{Advances in Neural Information
  Processing Systems (NeurIPS)}}.
\newblock


\bibitem[\protect\citeauthoryear{Zhang, Wang, Zhu, Xu, Wang, and Huang}{Zhang
  et~al\mbox{.}}{2018b}]%
        {zhang2018seq3seq}
\bibfield{author}{\bibinfo{person}{Xiaoyu Zhang}, \bibinfo{person}{Sheng Wang},
  \bibinfo{person}{Feiyun Zhu}, \bibinfo{person}{Zheng Xu},
  \bibinfo{person}{Yuhong Wang}, {and} \bibinfo{person}{Junzhou Huang}.}
  \bibinfo{year}{2018}\natexlab{b}.
\newblock \showarticletitle{Seq3seq fingerprint: towards end-to-end
  semi-supervised deep drug discovery}. In \bibinfo{booktitle}{\emph{ACM
  International Conference on Bioinformatics, Computational Biology, and Health
  Informatics (BCB)}}.
\newblock


\bibitem[\protect\citeauthoryear{Zhao, Ni, Yu, and Jiang}{Zhao
  et~al\mbox{.}}{2020}]%
        {zhao2020early}
\bibfield{author}{\bibinfo{person}{Tong Zhao}, \bibinfo{person}{Bo Ni},
  \bibinfo{person}{Wenhao Yu}, {and} \bibinfo{person}{Meng Jiang}.}
  \bibinfo{year}{2020}\natexlab{}.
\newblock \showarticletitle{Early anomaly detection by learning and forecasting
  behavior}. In \bibinfo{booktitle}{\emph{arXiv preprint arXiv:2010.10016}}.
\newblock


\bibitem[\protect\citeauthoryear{Zheng, Yan, Yang, and Xu}{Zheng
  et~al\mbox{.}}{2019}]%
        {zheng2019identifying}
\bibfield{author}{\bibinfo{person}{Shuangjia Zheng}, \bibinfo{person}{Xin Yan},
  \bibinfo{person}{Yuedong Yang}, {and} \bibinfo{person}{Jun Xu}.}
  \bibinfo{year}{2019}\natexlab{}.
\newblock \showarticletitle{Identifying structure--property relationships
  through SMILES syntax analysis with self-attention mechanism}. In
  \bibinfo{booktitle}{\emph{Journal of Chemical Information and Modeling}}.
\newblock


\bibitem[\protect\citeauthoryear{Zhou, Cao, Zhang, Trajcevski, Zhong, and
  Geng}{Zhou et~al\mbox{.}}{2019}]%
        {zhou2019meta}
\bibfield{author}{\bibinfo{person}{Fan Zhou}, \bibinfo{person}{Chengtai Cao},
  \bibinfo{person}{Kunpeng Zhang}, \bibinfo{person}{Goce Trajcevski},
  \bibinfo{person}{Ting Zhong}, {and} \bibinfo{person}{Ji Geng}.}
  \bibinfo{year}{2019}\natexlab{}.
\newblock \showarticletitle{Meta-GNN: On few-shot node classification in graph
  meta-learning}. In \bibinfo{booktitle}{\emph{International Conference on
  Information and Knowledge Management (CIKM)}}.
\newblock


\end{thebibliography}

\end{document}